\documentclass[a4paper]{cas-dc}

\usepackage[numbers]{natbib}
\usepackage{algorithm}
\usepackage{algorithmic}
\usepackage{subfigure}
\usepackage{enumitem}

\usepackage{caption}

\def\tsc#1{\csdef{#1}{\textsc{\lowercase{#1}}\xspace}}
\tsc{WGM}
\tsc{QE}

\begin{document}

\let\printorcid\relax

\shortauthors{Z.Liu et al.}  

\title [mode = title]{A Novel Underwater Image Enhancement and Improved Underwater Biological Detection Pipeline}

\author[1,2]{Zheng Liu}[type=editor,
        style=chinese]
\author[1,2]{Yaoming Zhuang}[type=editor,
        style=chinese]\cormark[1]\ead{zhuangyaoming524@163.com}
\author[1,2]{Pengrun Jia}[type=editor,
        style=chinese]  
\author[1]{Chengdong Wu}[type=editor,
        style=chinese]
\author[1]{Hongli Xu}[type=editor,
        style=chinese]
\author[3]{Zhanlin Liu}[type=editor,
        style=chinese] 





\affiliation[1]{organization={Northeastern University},
            addressline={Faculty of Robot Science and Engineering}, 
            city={Shenyang},
            postcode={110819}, 
            country={China}}
\affiliation[2]{organization={Northeastern University},
            addressline={College of Information Science and Engineering}, 
            city={Shenyang},
            postcode={110819}, 
            country={China}}
\affiliation[3]{organization={Warner Music Group},
            city={New York},
            postcode={10019}, 
            country={USA}}







\cortext[1]{Corresponding author}



\begin{abstract}
For aquaculture resource evaluation and ecological environment monitoring, automatic detection and identification of marine organisms is critical. However, due to the low quality of underwater images and the characteristics of underwater biological, a lack of abundant features may impede traditional hand-designed feature extraction approaches or CNN-based object detection algorithms, particularly in complex underwater environment. Therefore, the goal of this paper is to perform object detection in the underwater environment. This paper proposed a novel method for capturing feature information, which adds the convolutional block attention module (CBAM) to the YOLOv5 backbone. The interference of underwater creature characteristics on object characteristics is decreased, and the output of the backbone network to object information is enhanced. In addition, the self-adaptive global histogram stretching algorithm (SAGHS) is designed to eliminate the degradation problems such as low contrast and color loss caused by underwater environmental information to better restore image quality. Extensive experiments and comprehensive evaluation on the URPC2021 benchmark dataset demonstrate the effectiveness and adaptivity of our methods. Beyond that, this paper conducts an exhaustive analysis of the role of training data on performance.
\end{abstract}




\maketitle

\section{Introduction}\label{1}
The exploration of the aquatic environment has recently become popular due to the growing scarcity of natural resources and the growth of the global economy \cite{2021Lightweight}. Machine vision has been demonstrated to be a low-cost and dependable method with the benefits of noncontact monitoring, long-term steady operation, and a broad application range. Underwater object detection is pivotal in applications, such as underwater search and rescue operations, deep ocean exploration and archaeology, and sea life monitoring \citep{2020Real}. These applications necessitate effective and precise vision-based underwater sea analytics, including image enhancement, image quality assessment, and object detection methods. However, capturing underwater pictures with optical imaging systems has greater problems than in open-air conditions. More specifically, underwater images frequently suffer from degeneration due to color severe distortion, low contrast, non-uniform illumination, and noise from artificial lighting sources, which dramatically degrades image visibility and affects the detection accuracy of underwater object detection tasks \citep{2021Lightweight}. In recent years, underwater image enhancement technologies, work as a pre-processing operation to boost the detection accuracy by improving the visual quality of underwater images.

On the other hand, underwater object detection performances are associated with underwater biological living characteristics. Usually, because of differences in size or shape, as well as overlapping or occlusion of marine organisms, traditional hand-designed feature extraction methods cannot meet detection requirements in actual underwater scenes. Most of these studies emphasize extracting traditional low-level features such as color, texture, contour, and shape \citep{9416821}, leading to the disadvantages of traditional object detection methods such as poor recognition, low accuracy, and slow recognition. Directly benefiting from the deep learning methods, object detection has witnessed a great performance boost in recent years. However, due to the low quality of underwater imaging and complex underwater environment, the general object detection algorithm based on deep learning does not have a better detection effect on marine organisms.

The majority of the extant strategies consider underwater image enhancement and underwater object detection as two separate pipelines, whereas the underwater image enhancement is evaluated on the image quality assessment while the underwater object detection is evaluated on the detection accuracy. These two tasks have different optimization objectives, leading to different optimal solutions.

To validate our methods, we conducted experiments on the underwater robot professional contest (URPC2021) dataset. The experimental results prove the effectiveness of our proposed underwater image enhancement method and the improved underwater object detection algorithm. In conclusion, the primary contributions can be elaborated in the following three folds:
\begin{enumerate}
    \item To correct the bluish and greenish background and low contrast, an improved global histogram stretching method is proposed by dynamically adjusting the histogram stretching coefficient, for which a detailed linear function and framework are built.
    \item Fix the convolutional block attention module (CBAM) into the CSPDarknet53 to enhance the feature of small, overlapping, and occlusion objects. In particular, the CBAM mechanism is employed to improve the contrast between the object and the surrounding environment and refine redundant information produced by Focus Function.
    \item It is the first time using a simple and efficient connection between the attention mechanism and object detection algorithm. The CBAM module is added behind the Focus module of the backbone to reduce model burden as it as possible while ensuring the detection accuracy of the improved algorithm.
\end{enumerate}

The rest of the paper is organized as follows. Related works are elaborated in Section 2. Section 3 describes the materials and methods, mainly including the detail detection algorithm of the improved YOLOv5. In Section 4, the improved underwater model experiments and performance analysis are implemented. In the end, the conclusion is presented in Section 5.

\section{Related work}
\subsection{Underwater image enhancement methods (UIE)}

Underwater image enhancement is a necessary step in improving the visual quality of underwater images and is divided into three categories: model-free, physical model-based, and deep learning-based approaches. 

White balance \citep{Joost2007Edge}, Gray World theory \citep{2008A}, and histogram equalization \citep{1994Contrast} are examples of model-free enhancement methods that improve the visual quality of underwater images by directly adjusting the pixel value of images. Ancuti et al. suggested a multi-scale fusion underwater image enhancement method, which can be combined with fusion color correction and contrast enhancement to obtain high-quality images \citep{6247661}. Based on prior reasearch, Ancuti et al. proposed a weighted multi-scale fusion method for underwater image white balance, and restored the fading information and edge information of the original image through gamma variation and sharpening \citep{2017Color}. Fu et al. proposed a Retinex-based enhancement system that included color correction, layer decomposition, and underwater image enhancement in Lab color space \citep{7025927}. Zhang et al. extended retinex-based method by using bilateral and trilateral filters to enhance the three channels of underwater image in CIELab color space \citep{2017Underwater}. Because the physical process of underwater image deterioration is not taken into account, the model-free UIE approach may generate noise, artifacts, and color distortion, making it unsuitable for various types of underwater images.

Physical model-based methods regard underwater picture enhancement as an inverse image degradation problem, and these algorithms can provide a clear image by calculating transmission and background light based on Definition~\ref{definition1}. Because the underwater imaging model is similar to the atmospheric model on fog, the dehazing algorithms are used to enhancement underwater images. He et al. proposed a dehazing algorithm based on dark channel prior (DCP), which can effectively estimate the thickness of fog and obtain fog-free images \citep{Kaiming2011Single}. Based on DCP, Drew et al. proposed underwater dark channel prior by considering red light attenuation in water \citep{2016Underwater}. Peng et al. developed a generalized dark primary color prior (GDCP) for underwater image enhancement that included adaptive color correction in the image creation model \citep{Yan2018Generalization}. Galdran et al. proposed an improved Dark Channel Prior algorithm, which restores the lost contrast of underwater images by restoring colors related to red waves \citep{GALDRAN2015132}. Model-based approaches often need prior information, and the quality of improved images is dependent on precise parameter estimate.

Deep learning enhancement methods usually construct convolutional neural networks and train them using pairs of degraded underwater images and high-quality counterparts \citep{9245532}. Li et al. suggested an unsupervised generative adversance network (WaterGAN) that generates underwater images from aerial RGB-D pictures and then trains an underwater image recovery network using the synthesized training data \citep{2017WaterGAN}. To produce paired underwater image data sets, Fabbri et al. suggested an underwater color transfer model based on CycleGAN \citep{2017Unpaired} and built an underwater image recovery network using a gradient penalty technique \citep{2018Enhancing}. Ye et al. proposed an unsupervised adaptive network for joint learning, which can jointly estimate scene depth and correct color from underwater images  \citep{2020Deep}. Chen et al. proposed two perceptual enhancement cascade models, which used gradient strategy feedback information to enhance more prominent feature information on the image \citep{9245532}. Deep learning UIE approaches based on composite image training, in general, necessitate a large number of data sets. Because the quality of the composite image cannot be guaranteed, it cannot be applied to underwater situations.

\subsection {Attention mechanism}

Some research works have been presented in the literature for attention mechanisms. The attention model enables the network to extract information at crucial areas with reduced energy consumption, therefore enhancing CNN performance. Wang et al. proposed a Residual Attention Network based on the Attention mechanism, which can continuously extract a large amount of Attention information \citep{2017Residual}. Hu et al. proposed SENet, an architectural unit of "squeeze" and "excitation". This module enhances network expressiveness by modeling interdependencies between channels \citep{2018Squeeze}. Woo et al. proposed CBAM, a lightweight module that combines feature channel and feature space to refine features \citep{2018CBAM}. This method can achieve considerable performance improvement while keeping the overhead small. 

\subsection {Underwater object detection algorithm}

Deep learning-based object detection algorithms are currently divided into two categories: one-stage regression detectors and two-stage region generation detectors. One-stage detection methods mainly include YOLO series \citep{Redmon_2016_CVPR}, SSD \citep{10.1007/978-3-319-46448-0_2}, RetinaNet \citep{Lin_2017_ICCV}, and RefineDet \citep{Zhang_2018_CVPR}, which directly predict objects without region proposal. Two-stage detection methods mainly include RCNN \citep{Girshick_2014_CVPR}, Fast RCNN \citep{Girshick_2015_ICCV}, Faster RCNN \citep{NIPS2015_14bfa6bb}, Cascade RCNN \citep{8578742}, etc. Initially, these object detection methods were used to the natural environment on land. As deep learning technology advances, more and more object detection algorithms are being applied to difficult underwater environments. Li et al. used FasterRCNN to detect fish species and achieved outstanding detection performance \citep{7404464}. Li et al. employed a residual network to detect deep-sea plankton. The findings of the experiments reveal that deep residual networks generalise on plankton categorization \citep{7485476}. Cui et al. introduced a CNN-based fish detection system and optimized it by data augmentation, network simplification, and training process acceleration \citep{7761223}. Huang et al. presented three data augmentation approaches for underwater imaging to imitate the Marine illumination environment \citep{HUANG2019372}. Fan et al. suggested a cascade underwater detection framework with feature augmentation and anchoring refinement. This strategy can address the issue of imbalanced undersea samples \citep{10.1007/978-3-030-58565-5_17}. Zhao et al. designed a new composite backbone network to detect fish species by improving the residual network and used it to learn the change information of ocean scenes \citep{9416821}. However, little work and research has been done in the field of underwater object detection using YOLO.

The above related works analyze the research of underwater image enhancement, attention mechanism, and one-stage object detection algorithm. Based on the above analysis, the object detection algorithm under the terrestrial condition can be migrated and applied to the underwater condition. However, the complex underwater environment and underwater biological properties lead to low detection accuracy. In response to the above problems, we present an improved UIE method and an improved YOLOv5 object detection algorithm to handle underwater image degradation problems and low detection accuracy in complex underwater.

\section{Methodology}

In this paper, we first present the overview of our proposed the self-adaptive global histogram stretching algorithm (SAGHS) framework. Then we will demonstrate how to overcome underwater biological characteristics with the convolutional block attention module (CBAM). Next, our designed novel connection between CBAM and backbone is delineated, followed by the training and inference. We will detail our approach in the following two subsections.

\subsection {Self-adaptive histogram stretching algorithm}

Due to the complex underwater environment, underwater images often have visual distortion, such as low contrast, color distortion, and foggy effect. Before performing computer vision tasks, it is natural to consider image enhancement pre-processing methods. With image enhancement, the image quality is improved, and high-quality images facilitate object detection.  The main motivation and guideline for designing the network architecture of the proposed framework are addressed as follows.

\begin{figure*}[!ht]
  \begin{center}
  \includegraphics[width=6in]{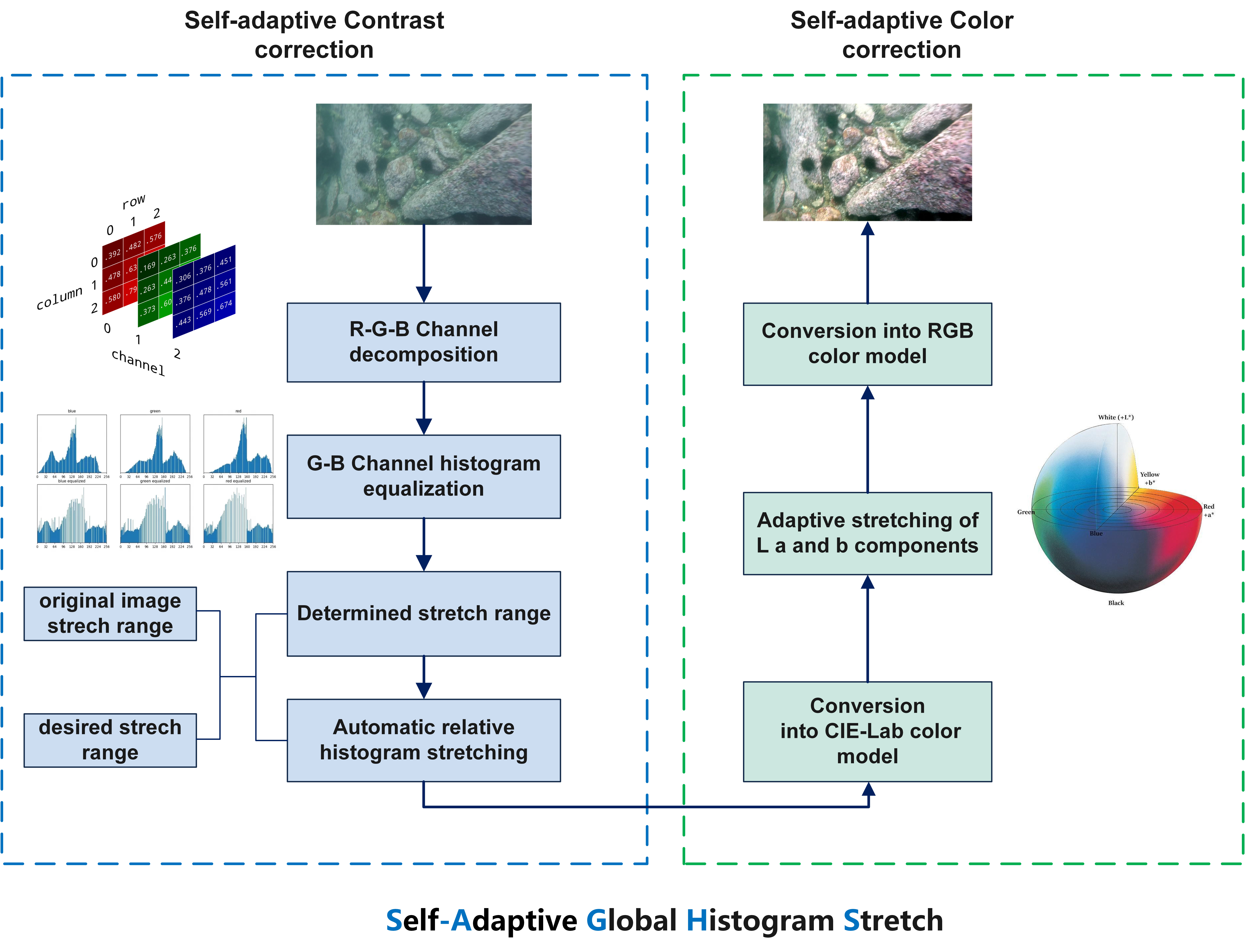}
  \caption{{The structure of Self-Adaptive Global Histogram Stretch. The self-adaptive contrast correction is the left part on the picture, and the self-adaptive color correction is the right part on the picture.}}\label{figure1}
  \end{center}
\end{figure*}

In this section, a two-stage self-adaptive global histogram stretching algorithm (SAGHS) based on structure decomposition and characteristics of underwater imaging is proposed, as shown in Figure~\ref{figure1}. The proposed SAGHA methods is a two-stage framework, including the self-adaptive contrast correction module and the self-adaptive color correction module. In the self-adaptive contrast correction module, the first step is to decompose the RGB channel and then apply color equalization and SAGHS to adjust the dynamic stretching range. In addition, the bilateral is used to eliminate noise after the above step. In the self-adaptive color correction module, the RGB image is converted to CIE-lab model, and then a simple linear histogram stretching is applied to adjust ‘L’ component. The adjusted CIE-Lab model converts into the RGB model finally. 
Basic principles and definition:
\newdefinition{definition}{Definition}
\begin{definition}
Simplified underwater optical imaging model [J-M model]:
\begin{equation}\label{definition1}
    I^C(x,y)=J^C(x,y)t^C(x,y)+B^C(1-t^C(x,y))
\end{equation}
\end{definition}
where $C\in\{R,G,B\}$. $I^C(x,y)$ represents the underwater image captured by the camera. $J^C(x,y)t^C(x,y)$ represents the part of the scene energy that directly decays. $J^C(x,y)$ represents scene radiance. $t^C(x,y)$ represents transmission map. $B^C(x,y)$ represents global background light. In water, $t^C(x,y)$ can be expressed as $e^{-\eta d(x,y)}$, Where $\eta$ is the attenuation coefficient, and $d(x,y)$ is the depth map between the scene and the camera.

\begin{definition}
Histogram slide stretching method [Iqbal]:
\begin{equation}\label{definition2}
    P_o=(P_i-c)\frac{(b-c)}{(a-c)}+a
\end{equation}
\end{definition}
where $P_i$ ,$P_o$ represents the input and output pixel values, respectively. $a,b$ represents the minimum and maximum value of the desired range, respectively. $c,d$ represents the lowest and highest pixel value currently present in underwater images.

\begin{definition}
Statistics of histogram distribution with a large of underwater images show that the distribution satisfies Rayleigh distribution, and its probability expression is
\begin{equation}\label{definition3}
    RD=\frac{x}{a^2}e^{\frac{-x^2}{2a^2}}
\end{equation}
\end{definition}
where the scale parameter $a$ of the distribution function is the mode, representing the peak value in each RGB histogram. When a channel presents a normal distribution, the median and mode are the same value.

\subsubsection {Self-adaptive contrast correction module}

The raw underwater images have low contrast and indistinct details due to light attenuation and dispersion. To properly adjust contrast and increase details, the first step is to decompose the RGB channels and then conduct color equalization for underwater images. After decomposing underwater image is inspired by the theory of the gray world hypothesis: for an ideal image, the average of the RGB channels on the image should be the "gray" K. Because red light attenuation in water is difficult to adjust by using simple color equalization, it will lead to the over-saturation of red light, so only according to the gray world assumption theory to correct the G and B channels, where half of the maximum value of each channel is selected to take K = 0.5 as the ‘gray value’. Next, the image channels are dynamically adjusted using the self-adaptive histogram stretching method. The specific function of dynamic adaptive stretching is documented in Algorithm 1.
\begin{algorithm*}[!h]
    \caption{Self-adaptive global histogram stretching}
    \label{alg:AOA}
    \renewcommand{\algorithmicrequire}{\textbf{Input:}}
    \renewcommand{\algorithmicensure}{\textbf{Output:}}
    \begin{algorithmic}[1]
        \REQUIRE original stretch range $I_{min}$, $I_{max}$, desired stretch range $O_{min}$, $O_{max}$, channel representation $\lambda$  
        \ENSURE Desired stretch minimum dynamic factor$\beta_\lambda$, Desired stretch maximum dynamic coefficient $\mu_\lambda$    
       
        \STATE  \textbf{Begin}
        \STATE  \textbf{function} stretchingRange($I_{min}$, $I_{max}$,$O_{min}$, $O_{max}$)
       
        \FOR{ $i,j \in [height,weight]$}
        \item
            initialize sort array
        \ENDFOR
        \STATE $I_{min},I_{max}$ is the pixel values of the 0.5\% and 99.5\% array respectively
        \STATE $O_{min} \in (0,I_{\lambda min})$
        \STATE $O_{min}=a_\lambda-\beta_\lambda\times\sigma_\lambda$
        \STATE $O_{max} \in (I_{\lambda max},255)$
        \STATE $O_{max}=I_{\lambda max}\div t_\lambda$
        \STATE $I_{max}=a_\lambda+\mu_\lambda\times\sigma_\lambda$
        \STATE \textbf{end function}
        \STATE \textbf{function} stretching($\beta_\lambda$,$\mu_\lambda$)
        \STATE Discuss the solution of $\beta_\lambda$,$\mu_\lambda$ to determine the stretch range and get the output pixel value $P_{out}$
        \STATE 
        $P_{out}=(P_{in}-I_{min})\frac{(O_{max}-O_{min})}{(I_{max}-I_{min})}+O_{min}$
        \RETURN $P_{out}$
    \end{algorithmic}
\end{algorithm*}

Such excessive stretching of certain color channels not only introduces noise that reduces the visibility of the image, but also introduces artifacts that cause color distortion and corrupt the original image details. According to the distribution pattern of RGB histogram in underwater images, the global histogram stretching equation(Definition~\ref{definition2}) is rewritten as Eq.~\ref{4}

\begin{equation}\label{4}
    P_{out}=(P_{in}-I_{min})\frac{(O_{max}-O_{min})}{(I_{max}-I_{min})}+O_{min}
\end{equation}
where $P_{in}$,$P_{out}$ represents the input and output pixel values, respectively. $I_{min},I_{max},O_{min},O_{max}$ represent the adaptive parameters of images before and after stretching respectively.

\textbf{Selection of stretching range $I_{min},I_{max}$:}
Generally, the stretching process is configured to follow the Rayleigh distribution (definition~\ref{definition3})and is restricted to a specific range. However, to reduce the effect of stretching due to extreme pixel points (e.g., noise, maxima, minima) suffered by the underwater image, this paper separates the upper and lower intensity values. In the proposed methods, the input intensity level is limited to 5\% of the minimum and maximum limitations. The restrictions are used to mitigate the effects of under and over-exposed areas in underwater images, which is shown in Eq.~\ref{5}.

\begin{equation}\label{5}
\begin{cases}
    I_{min}=\sigma_{st}[\sigma_{st.index}(a)\times0.5\%]\\ I_{max}=\sigma_{st}[-(\sigma_{len}-\sigma_{st.index}(a))\times0.5\%]
\end{cases}
\end{equation}
where $I_{min}$ and $I_{max}$ represent the minimum and maximum stretch values, respectively. $\sigma_{st}$ represents the ascending arrangement of the pixel set of each RGB channel. $\sigma_{st.index}(a)$ represents the index of the distribution pattern in the histogram. $\sigma_{len}$ represents the size of the image, and $\sigma_{st}[\cdot]$ represents the value in the index of the forward arraying pixel set.

\textbf{Selection of the desired range $O_{min},O_{max}$:}
The global histogram stretching period expects the stretching range to be [0,255], which brought excessive blue-green illumination to the underwater image. A simplified minimum desired stretch is obtained by calculating the standard deviation of the Rayleigh distribution.

\begin{equation}\label{6}
    O_{min}=a_{\lambda}-\beta_{\lambda}\times\sigma_{\lambda}
\end{equation}
Here, $a_{\lambda}$ is the mode of a channel, $\beta_{\lambda}$ is the dynamic minimum tensile coefficient, $\sigma_{\lambda}$ is the standard deviation of Rayleigh distribution, $\sigma_{\lambda}=0.655 a_{\lambda}$.

Next, the maximum desired stretch is determined based on the underwater imaging model (Definition 1) and the different attenuation levels exhibited by the light as it propagates through the water.

\begin{equation}\label{7}
    O_{max}=\frac{I_\lambda}{\kappa \times t_\lambda}=\frac{a_\lambda+\mu_\lambda\times\sigma_\lambda}{\kappa \times t_\lambda}
\end{equation}
The coefficient $\mu_{\lambda}$ is satisfied:

\begin{equation}\label{8}
    \frac{\kappa \times t_\lambda\times I_\lambda}{\sigma_\lambda} \leq \mu_\lambda+1.526 \leq \frac{\kappa \times t_\lambda\times 255}{\sigma_\lambda}
\end{equation}

In Eq.~\ref{6} and~\ref{8}, $\beta_{\lambda}$ and $\mu_{\lambda}$ have no solution or limited solution in integer filed. These adaptive parameters took into account both light transmission and the original image histogram distribution, and the image contrast is further rectified.

\subsubsection{Self-adaptive color correction module}

Following contrast correction in the RGB color model, the images are transformed into the CIE-Lab color model. In the CIE-Lab model, the L component represents the image brightness in the range of [0,100], indicating from brightest to darkest, $a$ denotes the component from green to red, while $b$ denotes the component from blue to yellow, and both $a, b$ values range in the range of [127, -128]. The $L$ component is applied with linear contrast stretching in this case, given by Eq.~\ref{9} within the range of [1\%,99\%]. The stretching of $a$ and $b$ component is defined as an S-model curve, given by Eq.~\ref{10}. The L-component stretching equation satisfies the linear sliding:

\begin{equation}\label{9}
    F_s(V)=\frac{V-min(V)}{max(V)-min(V)}
\end{equation}
$a$ and $b$ are defined as s-model curves:

\begin{equation}\label{10}
    O_x=I_x\times(\varphi^{1-{|\frac{I_x}{128}|}})
\end{equation}
where $I_x$,$O_x$ represents the input and output pixels, respectively. $x \in \{a,b\}$ represents $a, b$ components. $\varphi$ is the optimal experimental result value, and its value ranges from 1.2 to 2.0, where 1.3 is selected. This formula uses an exponential function as a redistribution coefficient. The closer the value to 0, the better the stretching effect is.

The color and luminance in an image are important parameters that make the image visible. The channels are composed after the $L$, $a$, and $b$ components have been stretched, and the image in the CIE-Lab color model is translated back into the RGB color model. After adaptive histogram stretching in the RGB color model and linear and nonlinear stretching adjustments in the CIE-Lab model, a clear image with high contrast, balancing, and saturation is finally obtained. Compared with the existing underwater image enhancement methods, this method can obtain better perceptual quality and less noise, thus improving the detection accuracy.

\subsection{The Convolutional block attention module mechanism}

\begin{figure}[!ht]
  \begin{center}
  \includegraphics[width=3.3in]{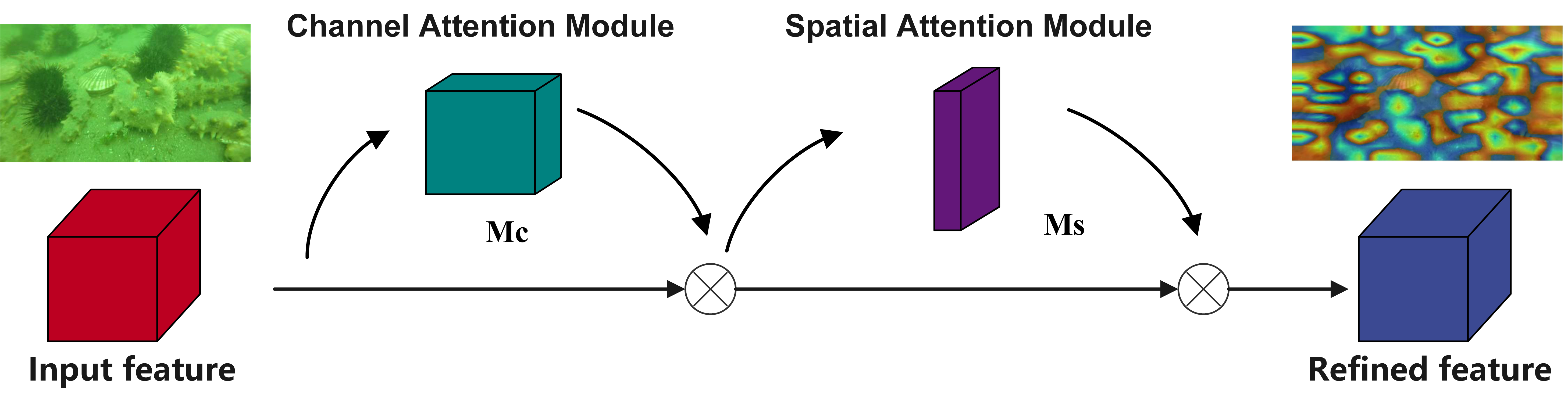}
  \caption{{The convolutional block attention module model structure. This module has two sequential sub-modules: channel attention module and spatial attention module. The intermediate feature map is adaptively refined through CBAM.}}\label{cbam}
  \end{center}
\end{figure}

The convolutional block attention module (CBAM) focuses on the important feature of the network while suppressing the unimportant information, effectively improving the performance of the CNN. Given the difficulty that marine organism objects are tiny and the features are not evident from the background, it is difficult for the model to extract and conserve the feature. The channel attention (CA) map controls the inter-channel relationships of features, while the spatial attention (SA) map is used to exploit the internal spatial relationships of features. The combined use of CA and SA can capture the dependency between attention at a fine-grained level and generalize well to enhance feature expression and improve detection accuracy during feature extraction. The CBAM structure is shown in Figure~\ref{cbam} and the formula is shown in Eq.~\ref{11}

\begin{equation}\label{11}
\begin{cases}
    F_1 = M_c(F)\otimes F\\F_2 = M_s(F_1)\otimes F_1
\end{cases}
\end{equation}

\subsubsection {Channel Attention module}


\begin{figure}[!ht]
  \begin{center}
  \includegraphics[width=3.3in]{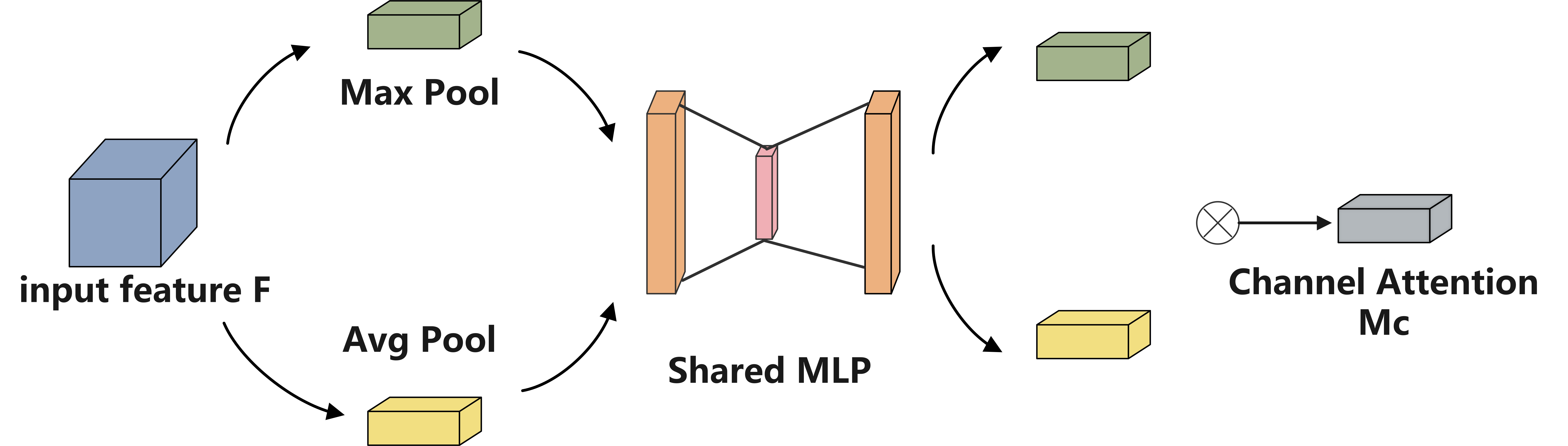}
  \caption{{Channel attention module structure. The channel module utilizes both max pooling outputs and average pooling outputs with a share network.}}\label{cam}
  \end{center}
\end{figure}
The structure is shown in Figure~\ref{cam}. The channel attention module compresses the spatial dimensions of feature maps using average-pooling and max-pooling, respectively. Max pooling takes into account just the biggest element, ignores smaller components in the pooling zone, and preserves more picture texture information. To retain more images background information, average pooling computes the average of all components in the pooling zone. The channel attention module uses both average and max pooling to aggregate the spatial information of a feature and delivers two different spatial context descriptors: $F^C_{avg}$ and $F^C_{max}$. To build a distinct channel attention map $M_c$, two descriptors are transmitted to a shared multilayer perception (MLP). The channel attention mechanism allows modeling the importance of individual feature channels and then enhancing or suppressing different channels for specific tasks. In this paper, we increase the weights of channels that contribute a lot to the detection accuracy and decrease the weights of channels that do not contribute much to the detection accuracy which ultimately improves the detection accuracy of the network.
In the channel attention module, the calculation formula of the weight coefficient matrix $M_c(F)$ is 
\begin{equation}\label{12}
\begin{split}
    M_c(F)=\sigma(W_1(W_0(F^c_{avg}))+W_1(W_0(F^c_{max})))\\
\begin{cases}
    MLP(AvgPool(F))=W_1(W_0(F^c_{avg}))\\
    MLP(MaxPool(F))=W_1(W_0(F^c_{max}))
\end{cases}   
\end{split}
\end{equation}
where $\sigma$ represents the sigmoid activation function, $W_0$,$W_1$ represents the weight of MLP, $W_0\in R^{C/r\times C}$, $W_1 \in R^{C\times C/r}$, and r=16 represents the reduction ratio.

\subsubsection{Spatial attention module}


\begin{figure}[!ht]
  \begin{center}
  \includegraphics[width=3.3in]{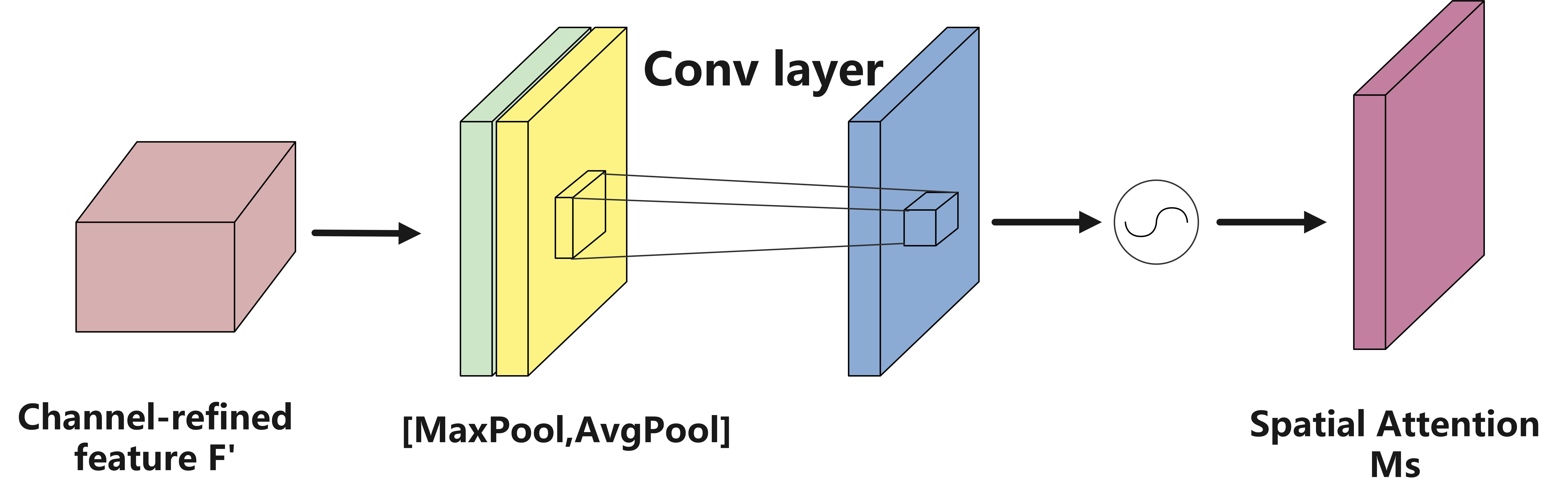}
  \caption{{Spatial attention module structure. The spatial module utilizes similar two outputs that are pooled along the channel axis and forward them to a convolution layer.}}\label{sam}
  \end{center}
\end{figure}

The spatial attention module differs from the channel attention block in that the spatial attention module focuses more on distinguishing the location of features and complementing the channel attention mechanism. The structure is shown in Figure~\ref{sam}. The spatial attention module applies average pooling and max pooling along the channel dimensions, respectively, and two feature maps $F^s_{avg}$and $F^s_{max}$ are combined after the max pooling and average pooling. The number of channels is reduced to 1 by dimensionality reduction filtering of 7×7 convolution kernel. Finally, the spatial attention feature map is obtained by the sigmoid activation function.

\begin{equation}\label{13}
\begin{split}
    M_c(F)=\sigma(f^{7\times7}([F^s_{avg};F^s_{max}]))\\
\begin{cases}
    F^s_{avg}=AvgPool(F)\\
    F^s_{max}=MaxPool(F)
\end{cases}   
\end{split}
\end{equation}

The spatial attention module utilizes global contextual information. It exploits the spatial attention mechanism to selectively capture spatial interdependence between sites in order to produce a typical contribution of points in the spatial dimension and extract more robust marine biological features. The interrelationships between channel mappings are explored to model the importance of each feature channel and enhance the feature representation of organisms. The attention mechanism is added to the backbone feature extraction network, which mainly uses the attention module to focus on the actual content information of the detected target. It is effective for the output results of the backbone feature extraction network. 

\subsection{Enhanced YOLOv5 Network}

\begin{figure*}[!ht]
  \begin{center}
  \includegraphics[width=6in]{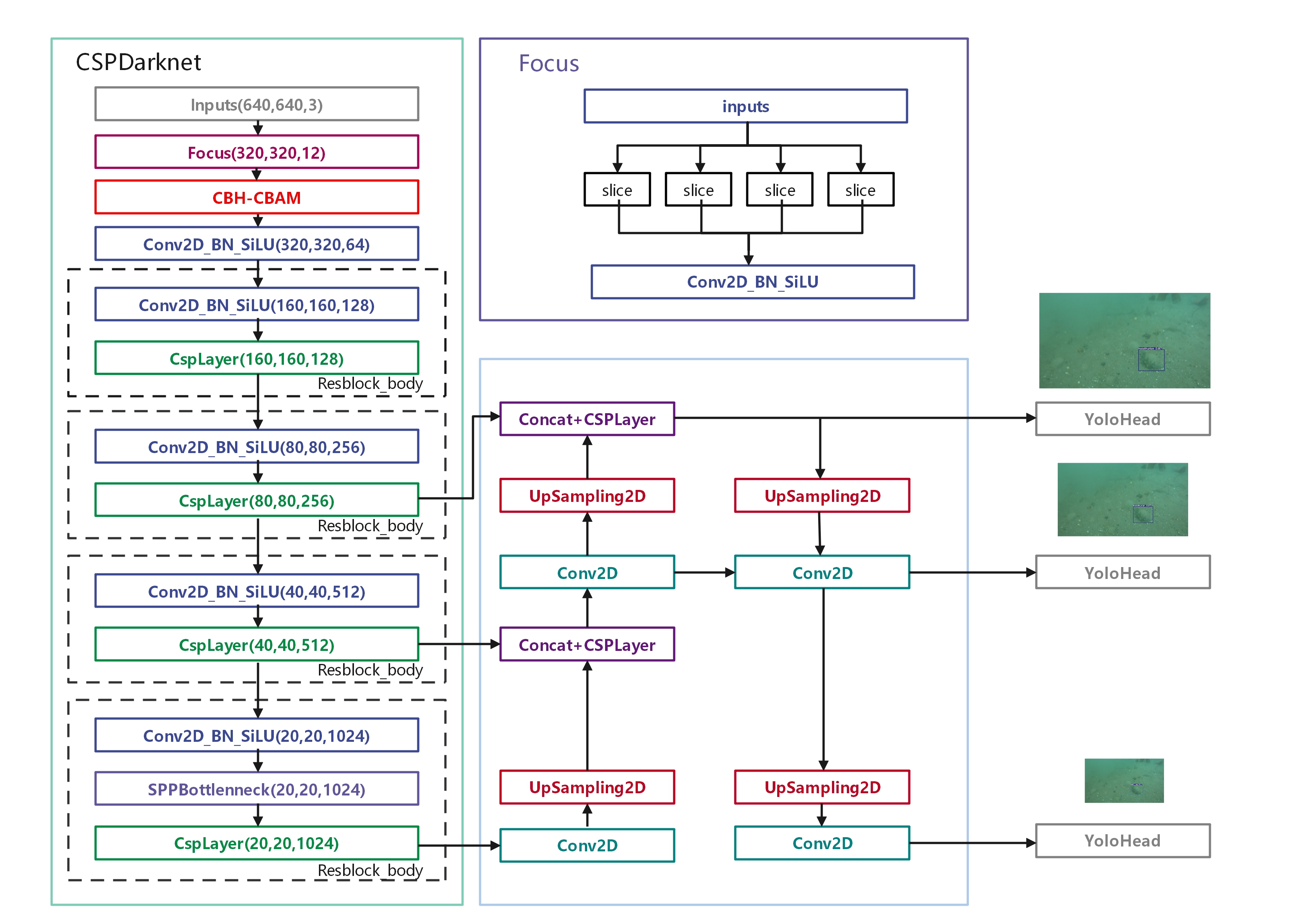}
  \caption{{Improved YOLOv5 network structure combined with CBAM. For each layer, [W, H, C] means the size, including the width (W), the height (H), and the number of channels (C), of the output feature from this layer.}}\label{yolo}
  \end{center}
\end{figure*}

The YOLOv5 network has fine-tuned in this part. Based on the original YOLOv5 detection model, the CBAM mechanism is added to the backbone feature extraction network to construct the CBAM-YOLOv5 detection algorithm. The CBAM mechanism helps the convolutional feature network model to learn the feature weights of different regions and aim to the characteristics of denseness, mutual occlusion, and multiple small objects of marine organism. CBAM is a lightweight plug-and-play module, which can be integrated into a convolutional neural network for end-to-end training. This paper designed a simple and effective connection to improve the performance of detection accuracy with the cost of a slight increase in computation. The improved network structure is shown in Figure~\ref{yolo}.


\subsubsection{YOLOv5 object detection algorithm}
The one-stage detection model of the YOLO series mainly consists of three components: Backbone, Neck, and Head. The YOLOv5 model is fast and flexible compared to other versions of the YOLO series. The input image is fed into the backbone feature extraction network for slicing operation (Focus) to reduce the height and width of the image, and the cut image height and width are integrated through the Concat function to increase the number of input channels for feature extraction using the convolution module. Then the extracted feature maps were followed by three sets of simplified CSP modules, CONV operations, and used the SPP module to improve the detection accuracy of the model. The features were extracted by four max pooling operations aggregated by Concat. In the YOLOv5 structure, Neck and Head are not distinguished. The path aggregation network and detection section are included directly in the detection head. The second structural module of CSP is used in the path aggregation part to reduce the number of parameters. The path aggregation network structure improves the detection effect of small objects by integrating high and low-level features. A bottom-up path expansion structure is introduced to segment the network using its shallow features. The object detection task is pixel-level classification, with exterior features such as edge forms being prominent. The new bottom-up augmentation allows feature mapping at the top layer to benefit from the extensive position information provided by the bottom layer, thus enhancing large object recognition. The detection part of the head uses a multi-scale feature map for detection, using large images to detect small targets and small images to detect large targets. In the detection head of YOLOv5, candidate boxes are generated on feature maps of three different scales, and the bounding boxes are filtered by weighted NMS, the object classification and boxes regression are output.

\subsubsection{Improved YOLOv5 backbone network}

\begin{figure*}[!ht]
  \begin{center}
  \includegraphics[width=6in]{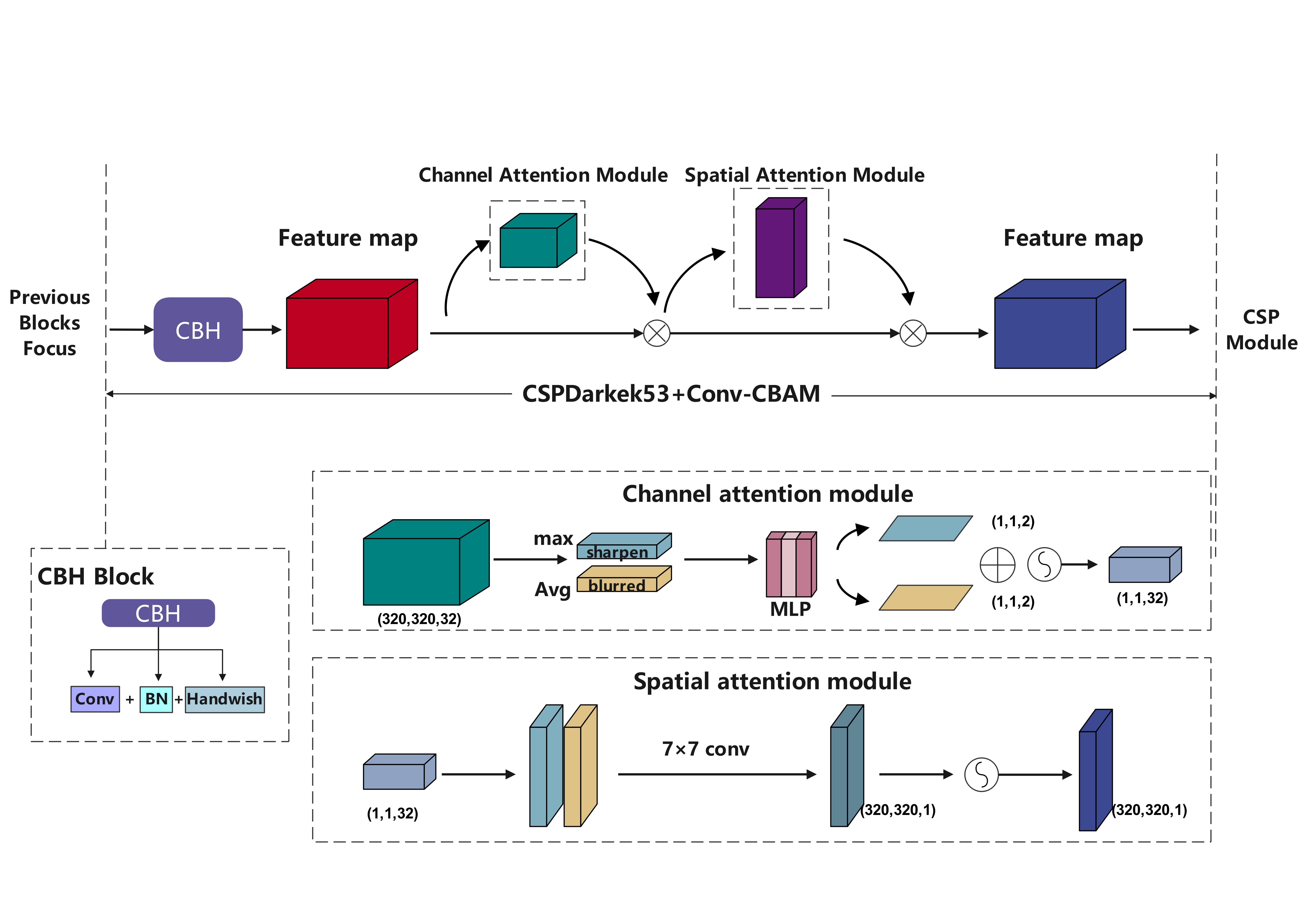}
  \caption{{CBAM integrated with a convolutional block in CSPDarknet53. This figure shows the exact position of our module when integrated within a CSPDarknet53.}}\label{improved}
  \end{center}
\end{figure*}

 Although the channel attention mechanism and spatial attention mechanism module are identical in principle, both focusing on local information to improve detection accuracy, the structure of the attention mechanism varies from location to location when being applied to different areas. In this paper, we find a more concise and efficient solution by adding the convolution block attention mechanism after the first convolution block in the backbone, as shown in Figure~\ref{improved}. Through experimental analysis and literature reading, we found that placing the convolutional block attention module at the beginning of the backbone network can effectively reduce the interference brought by the complex water environment to underwater biological detection.


In figure~\ref{improved}, it is assumed that the input image size of the YOLOv5s model is 640×640×3, and 320×320×32 in CBAM after the Focus slice. After entering the channel attention module, the sharpened 1×1×32 feature map was obtained after global max pooling, and the blurred 1×1×32 feature map was obtained after global average pooling. The feature maps obtained by parallel pooling lose less information and have strong localization ability. Then the feature maps entered MLP module too reduce the dimension to 1×1×2. The nonlinear data after MLP shared were classified and the dimensionality reduction coefficient was 16, then the dimension of feature maps was increased to 1×1×32. The MLP output features are subjected to element-wise, and after sigmoid function activate, the channel attention feature map Mc of size 1×1×32 was generated. The input feature map F is multiplied element-wise with the channel attention feature to get an output size of 320×320×32. Next, the output feature map is treated as input to the spatial attention module. In the spatial attention module, the designed is symmetric with the channel attention. Through the channel-based global max pooling and global average pooling, respectively, two feature maps with the size of 320×320×1 was obtained. The channels of the two feature maps are merged into a feature map with the size of 320×320×2 by Concat Function, and then the channels' dimensionality was reduced to 1 by 7×7 convolution. Finally, the Sigmoid activation function is used to get a spatial attention map with a size of 320×320×1. The input of the spatial attention module is multiplied by the spatial attention to obtain an output feature map with a size of 320×320×32. The output feature diagram of CBAM is consistent with the input feature diagram.
\section {Experimental Configuration}

\subsection {Dataset}

\begin{figure}[!ht]
  \begin{center}
  \includegraphics[width=3.3in]{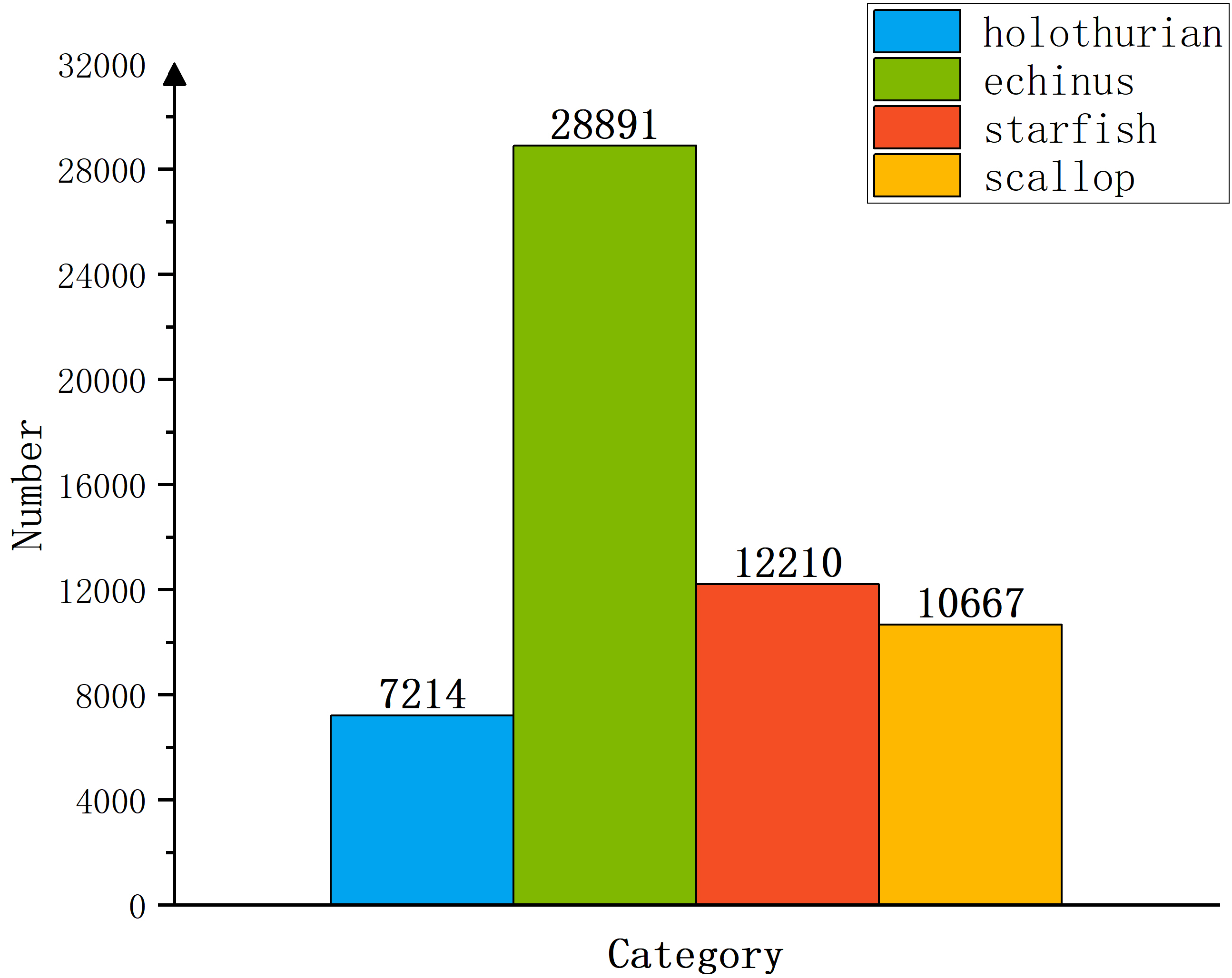}
  \caption{{The statistics of the number of categories in URPC2021 training dataset.}}\label{zhu}
  \end{center}
\end{figure}

In this study, the Underwater Robot Professional Contest 2021 (URPC2021 Dalian) benchmark dataset is used for training. This benchmark dataset was created primarily to give resources for evaluating underwater domain detection algorithms in picture and video sequences. The images of this benchmark dataset are obtained by frame rate interception of the video captured by the underwater robot ROV in the natural environment. The URPC2021 dataset provides 8200 underwater images and box-level annotation. There are four categories in this dataset: holothurian, echinus, starfish, and scallops. We randomly divided all the pictures into training set and validation set according to the ratio of 0.85:0.15. Beyond that, we made statistics on the number distribution of objects for each category in Figure~\ref{zhu}. 

\begin{figure}[!ht]
    \centering
    \subfigure[]{%
    \includegraphics[width=1.5in]{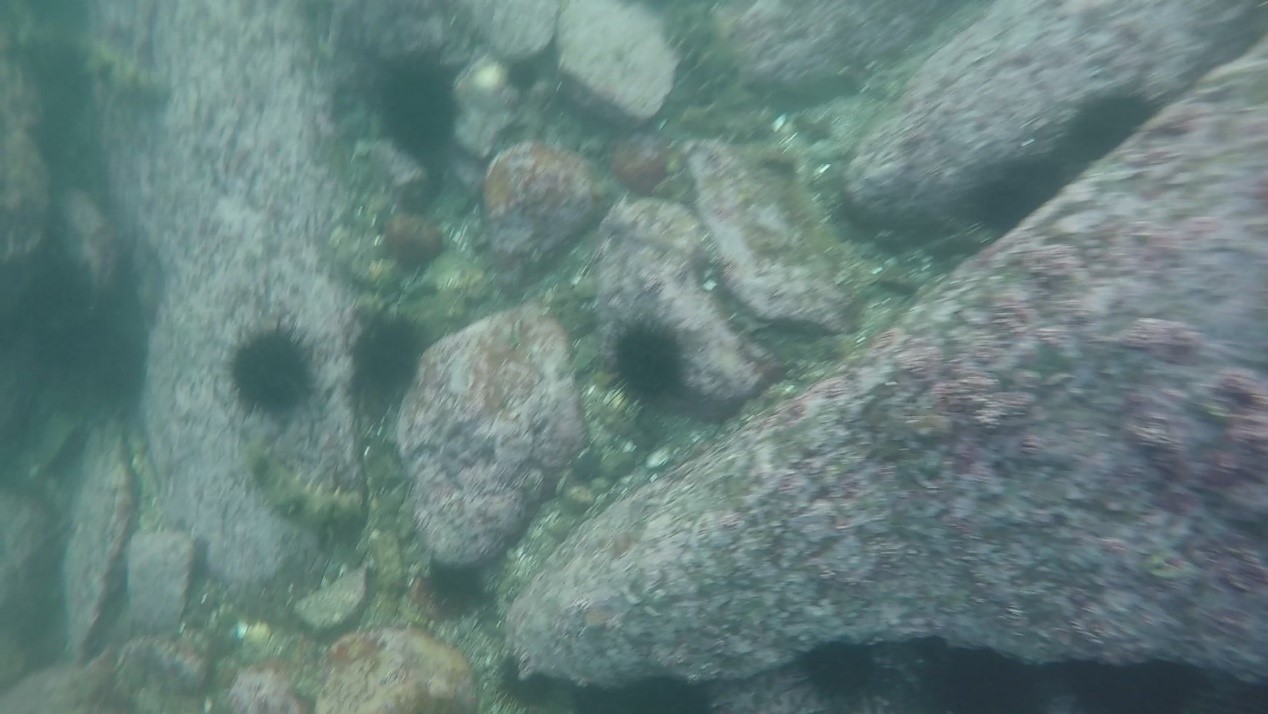}
    \label{fig:subfigure1}}\subfigure[]{%
    \includegraphics[width=1.5in]{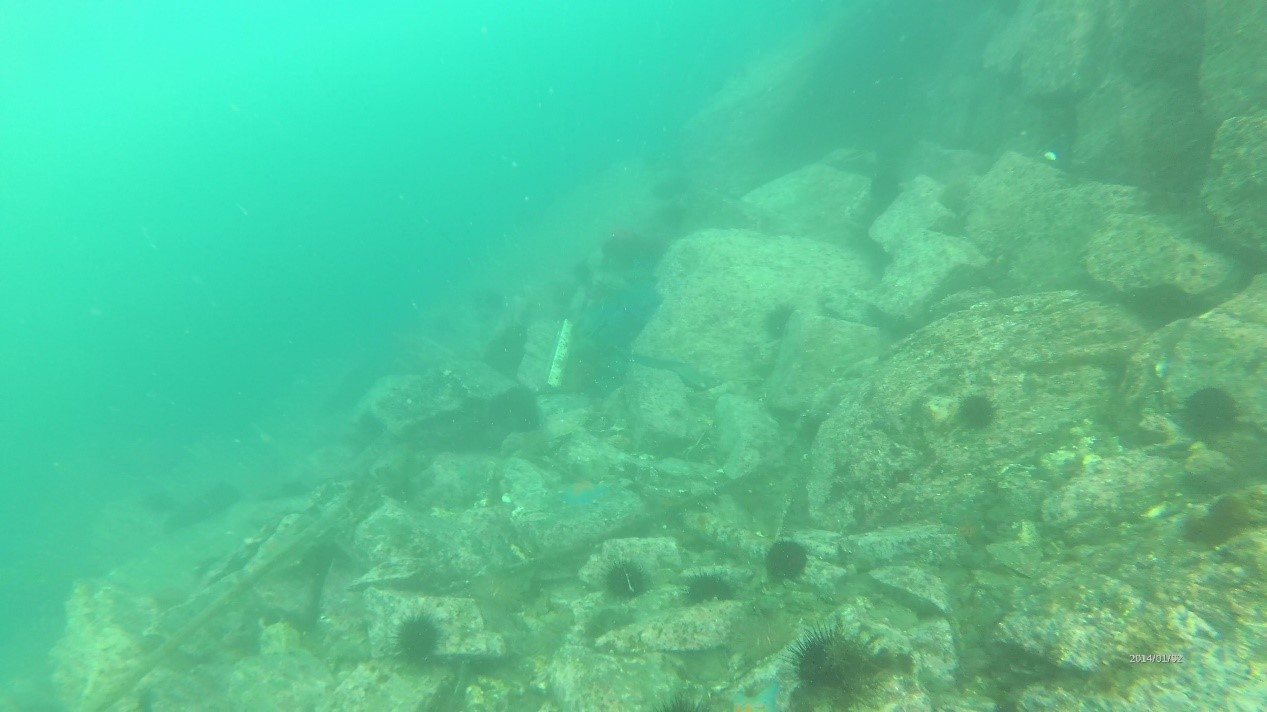}
    \label{fig:subfigure2}}
    \subfigure[]{%
    \includegraphics[width=1.5in]{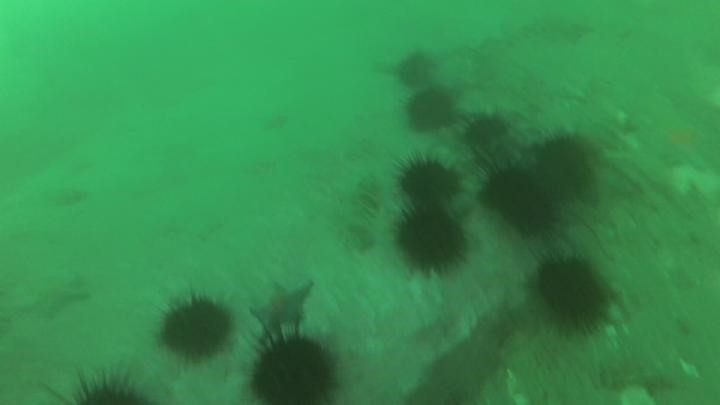}
    \label{fig:subfigure3}}\subfigure[]{%
    \includegraphics[width=1.5in]{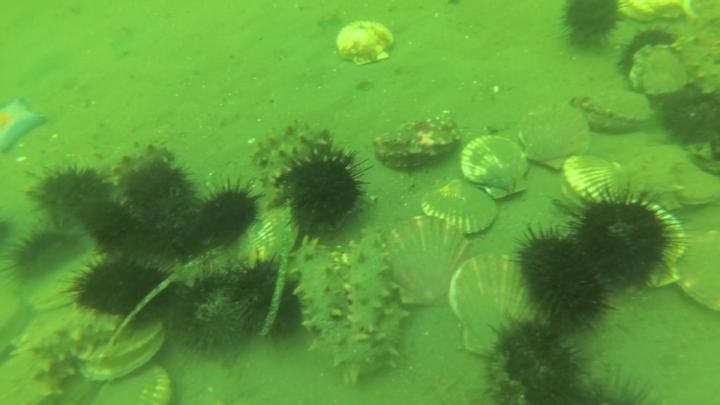}
    \label{fig:subfigure4}}
\caption{ Four typical scenarios in the URPC2021 dataset are difficult to recognize. (a) Low-resolution underwater images (b) Motion blur (c) Color cast and low contrast (d) Objects density and small}
\label{fig:figure}
\end{figure}

The complex environment of this scenario presents a substantial difficulty for marine organism identification, as evidenced by the four elements depicted in Figure~\ref{fig:figure}. The following main obstacles for marine organism identification are posed by the complex underwater environment:
\begin{itemize}
\item Underwater images with low resolution. The textural feature information of aquatic organisms is lost in low-quality images, making it more difficult to recognize creatures with comparable features.
\item Motion blur. Since the dataset is obtained from video clipping, motion blur is inevitable when sampling robot movement. In low light conditions, there is little difference in the morphology of underwater creatures.
\item Underwater environment color cast and low contrast. Affected by the propagation properties of light underwater, underwater image datasets are mostly blue-green backgrounds with low contrast, which makes creatures such as scallops easy to confuse the background.
\item The target organisms are small and gather to block each other. The density of underwater creatures is high and mutual, resulting in a serious loss of texture information of occluded creatures.
\end{itemize}  

\subsection {Experiments details}

The Yolov5 model in this paper can adapt to image scaling, select 640×640 size as the input of the image, and obtain the feature map of equal size as the detection scale. The best parameters are found that the learning rate of 0.01 can achieve convergence faster, and the training speed is faster when the batch size is 16. The hardware environment and software version training platform parameters used are an Inter(R) Xeon(R) silver 4208 CPU@ 2.10Hz, an NVIDIA RTX 2080super(8G) GPU, the operating system is Ubuntu18.04, and the Pytorch environment of CUDA10.2, torch=1.7.1. For all comparision models, the initial learning rate is set to 0.01. The momentum =0.937 and weight\_decay=0.0005. The number of training rounds is unified to 100 epochs.

\subsection{Evaluation indicators}

To evaluate the performance of our proposed architecture for underwater small creature detection, we consider the performance of the model by analyzing the experimental results for precision, recall , mean average precision (mAP), F1 score, frames per second (FPS).

\textbf{Precision} refers to the ratio of correctly predicted positive samples to all indicated positive examples, and high precision represents a high confidence level for this class of organisms.

\textbf{Recall} refers to the percentage of correctly predicted positive samples out of all positive samples. For underwater small organism detection, a high recall means that the algorithm can detect a certain type of organism in the dataset completely.

\textbf{F1 score} Since precision and recall are often mutually exclusive. The F1 score is used as an overall measure of the quality of the algorithm being designed.

\textbf{mAP} is the average AP value across multiple categories. We plotted the P-R curve with precision as the vertical coordinate and recall as the horizontal coordinate. AP is the area between the curve and the axes. 

\textbf{Frames per second (FPS)} For detection tasks with real-time requirements, FPS is an important measure of the model's real-time performance, and the shorter the time, the better, provided that the quality of detection is guaranteed.

\subsection{Results and discussion}

\subsubsection {Image preprocessing experiments}

\begin{figure*}[!ht]
	
 \begin{minipage}{0.32\linewidth}
 	\vspace{3pt}
 	\centerline{\includegraphics[width=\textwidth]{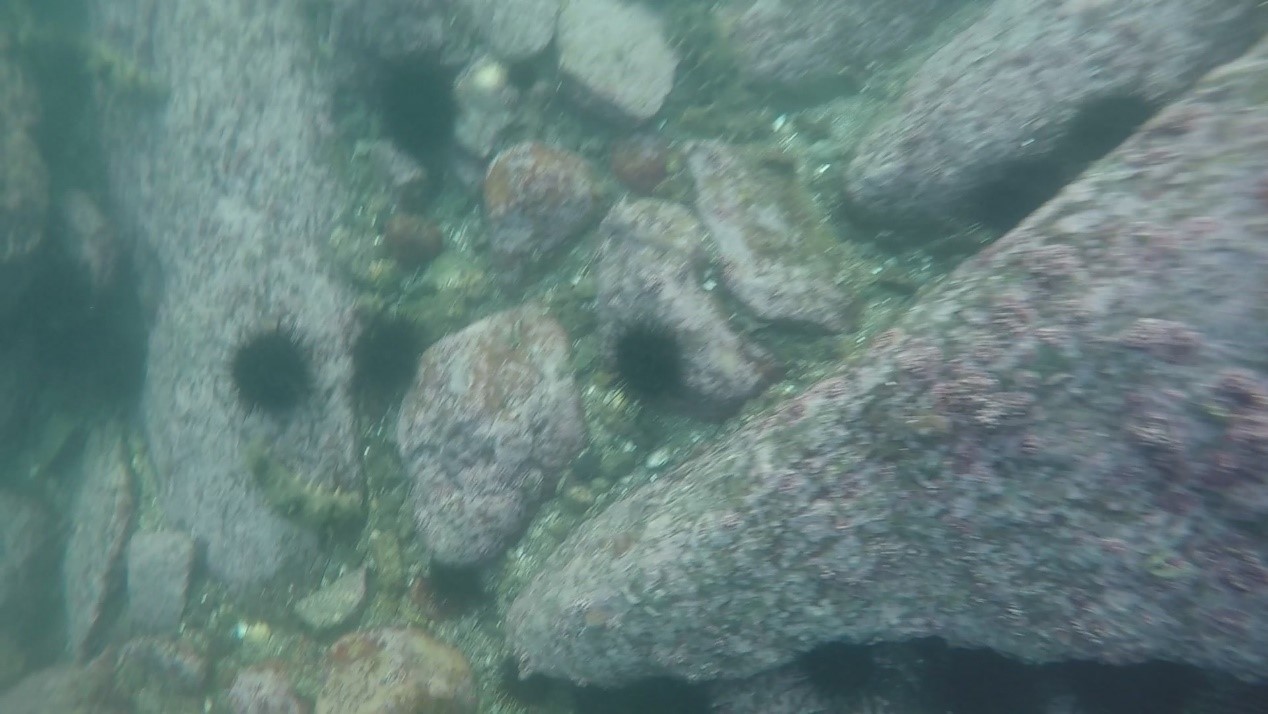}}
 	\vspace{3pt}
 	\centerline{\includegraphics[width=\textwidth]{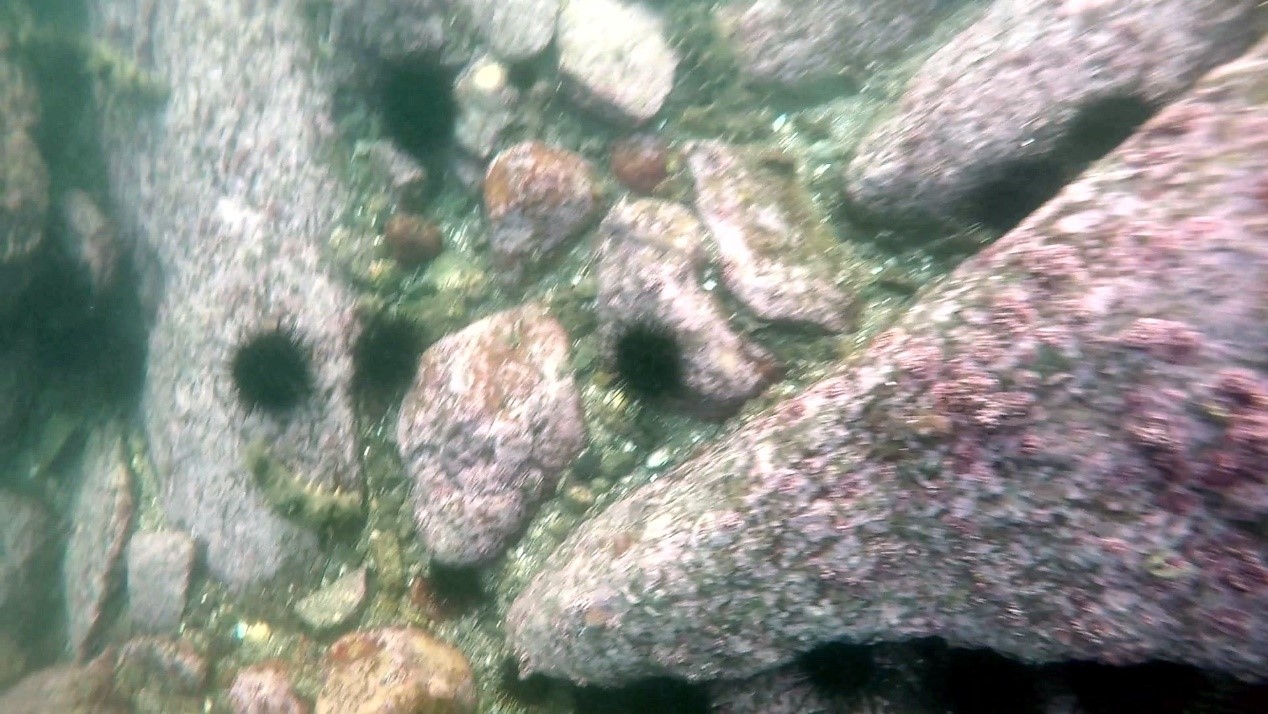}}
 	\vspace{3pt}
 	\centerline{(a)}
 \end{minipage}
 \begin{minipage}{0.32\linewidth}
	\vspace{3pt}
	\centerline{\includegraphics[width=\textwidth]{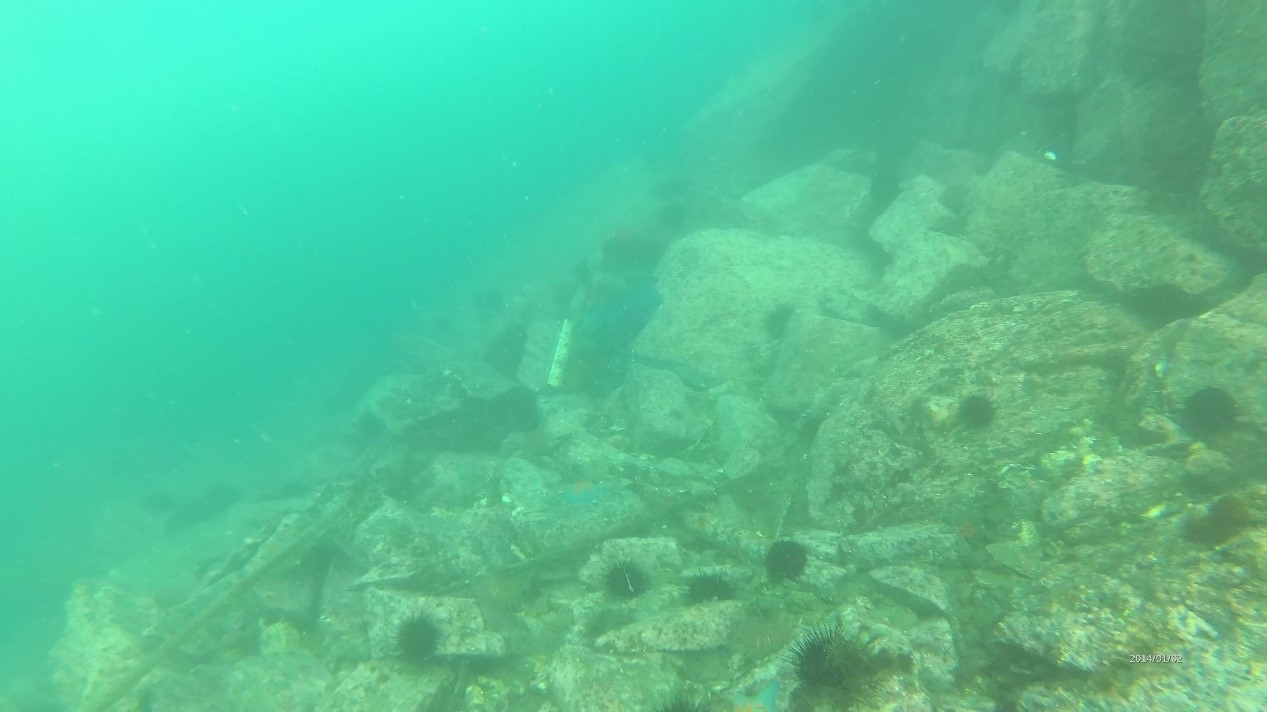}}
	\vspace{3pt}
	\centerline{\includegraphics[width=\textwidth]{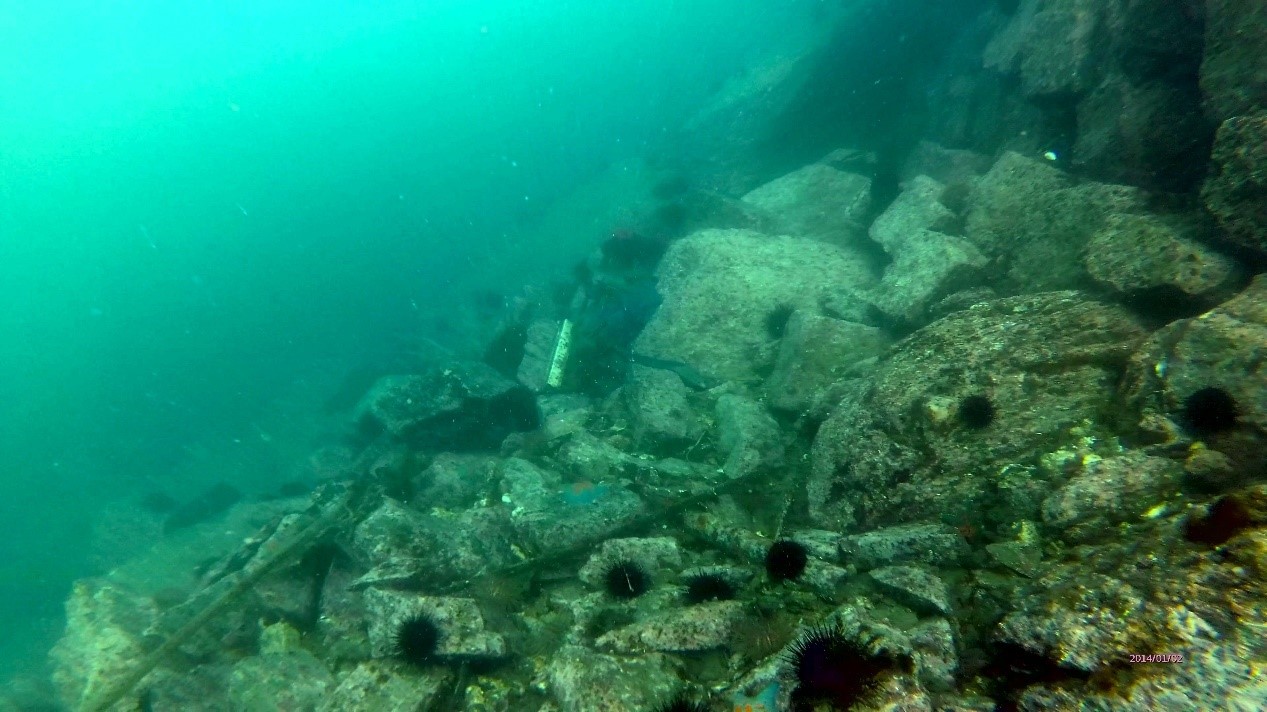}}
	\vspace{3pt}
	\centerline{(b)}
\end{minipage}
\begin{minipage}{0.32\linewidth}
	\vspace{3pt}
	\centerline{\includegraphics[width=\textwidth]{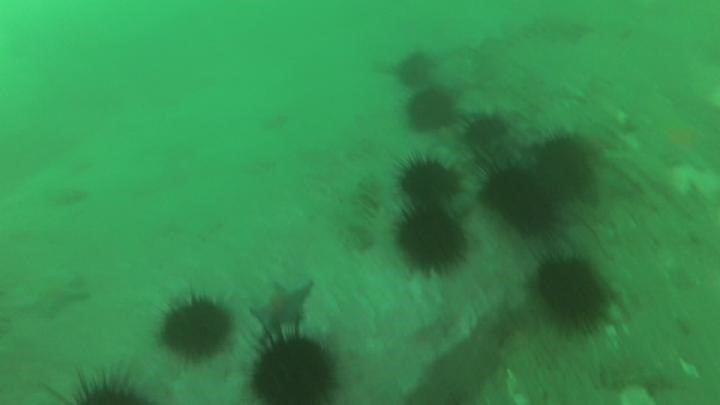}}
	\vspace{3pt}
	\centerline{\includegraphics[width=\textwidth]{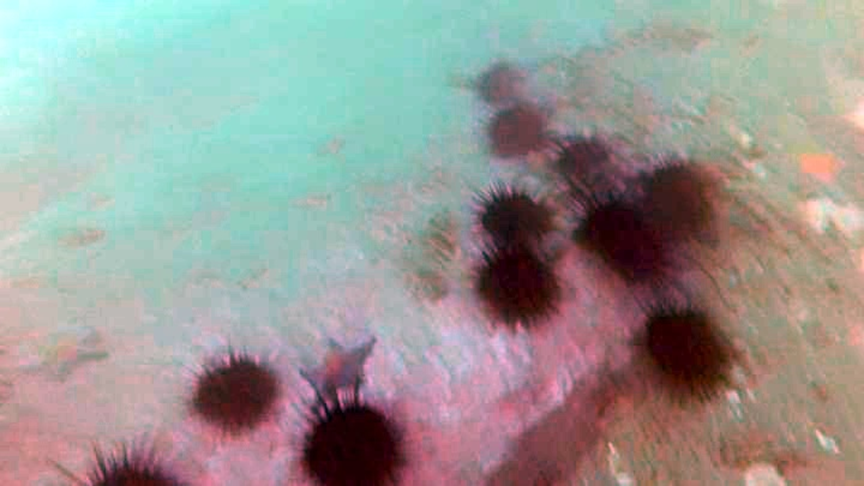}}
	\vspace{3pt}
	\centerline{(c)}
\end{minipage}
\caption{ Enhancement results of the underwater images. The top row is RAWS, and the bottom row is the results of the proposed SAGHS method. (a) Low contrast (b) Bluish background (c) greenish background}
\label{comparise}
\end{figure*}

\begin{figure*}[!ht]
	
 \begin{minipage}{0.24\linewidth}
 	\vspace{3pt}
 	\centerline{\includegraphics[width=\textwidth]{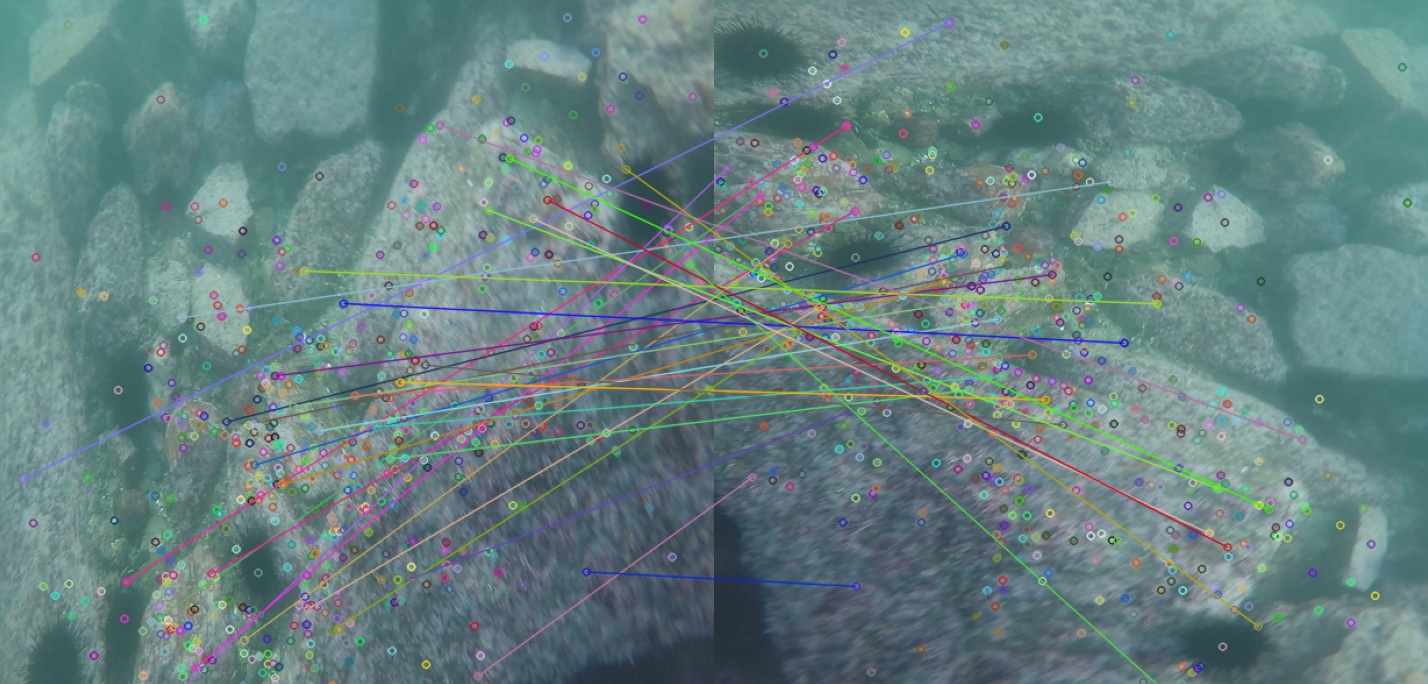}}
 	\vspace{3pt}
 	\centerline{\includegraphics[width=\textwidth]{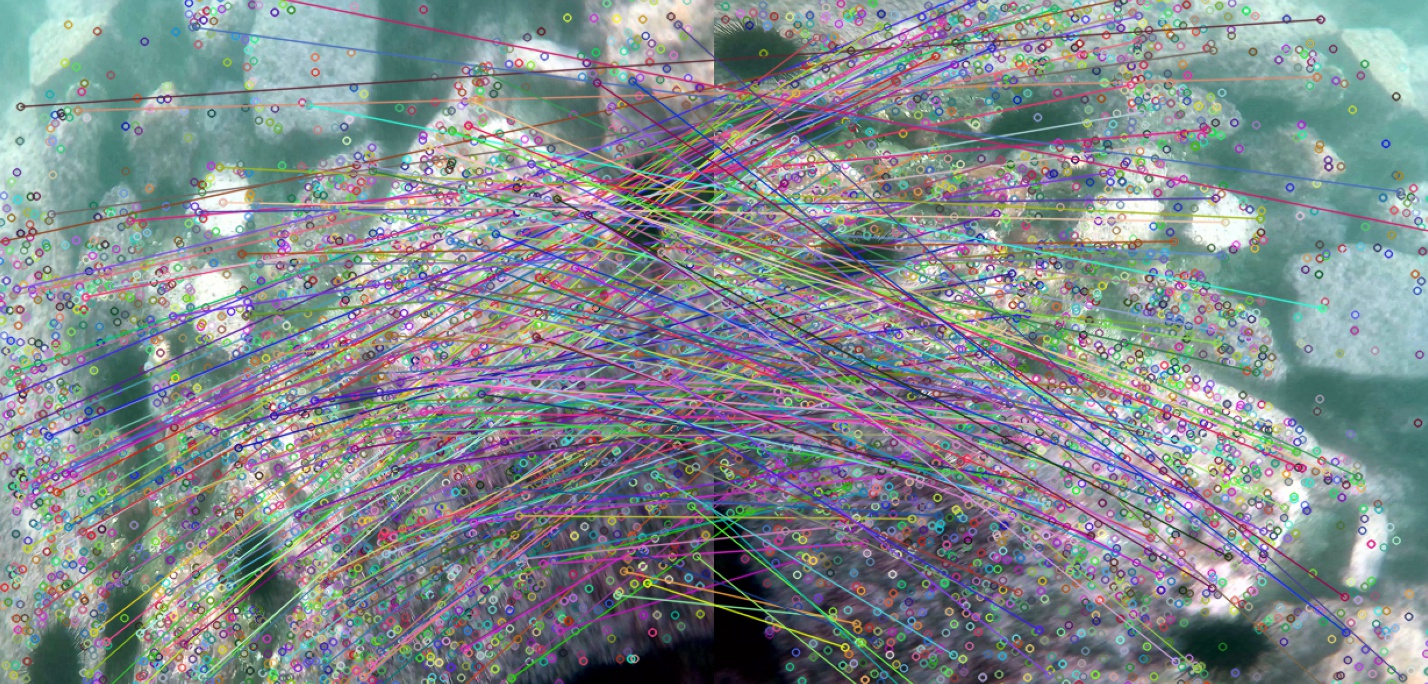}}
 	\vspace{3pt}
  	\centerline{(a)}
 \end{minipage}
 \begin{minipage}{0.24\linewidth}
	\vspace{3pt}
	\centerline{\includegraphics[width=\textwidth]{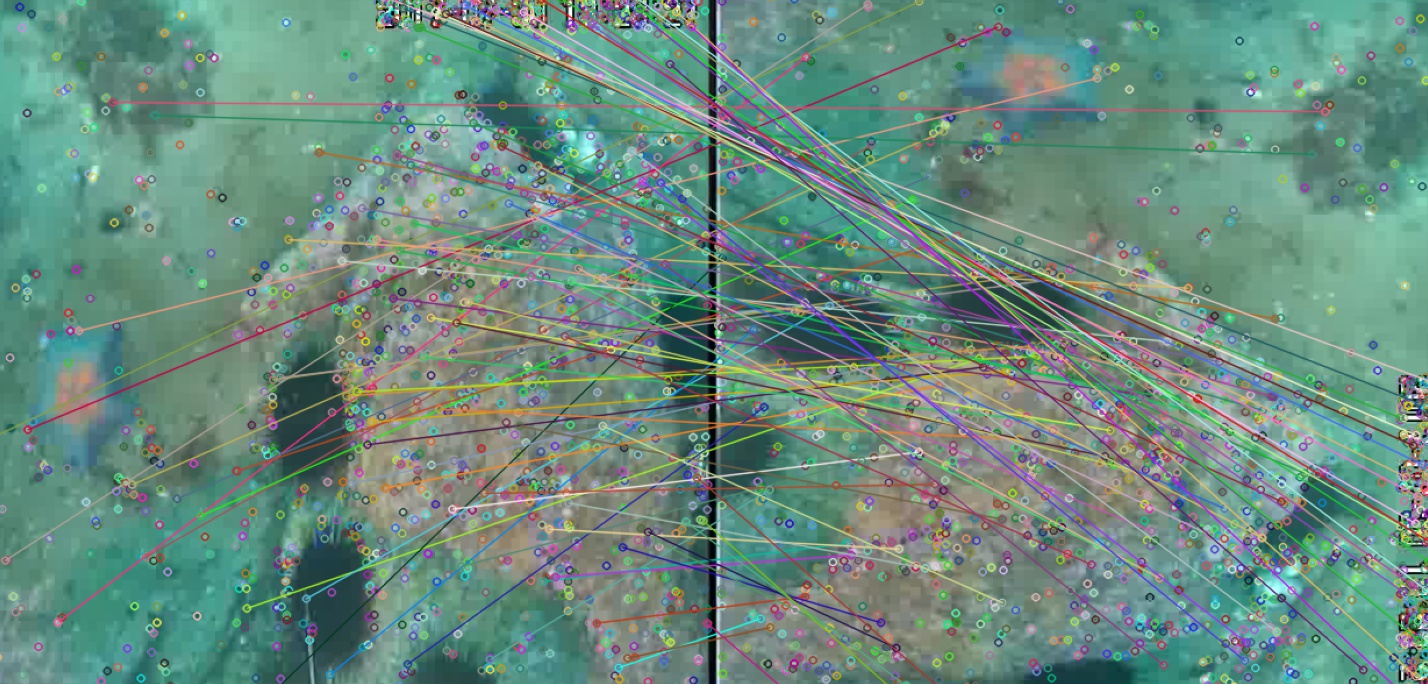}}
	\vspace{3pt}
	\centerline{\includegraphics[width=\textwidth]{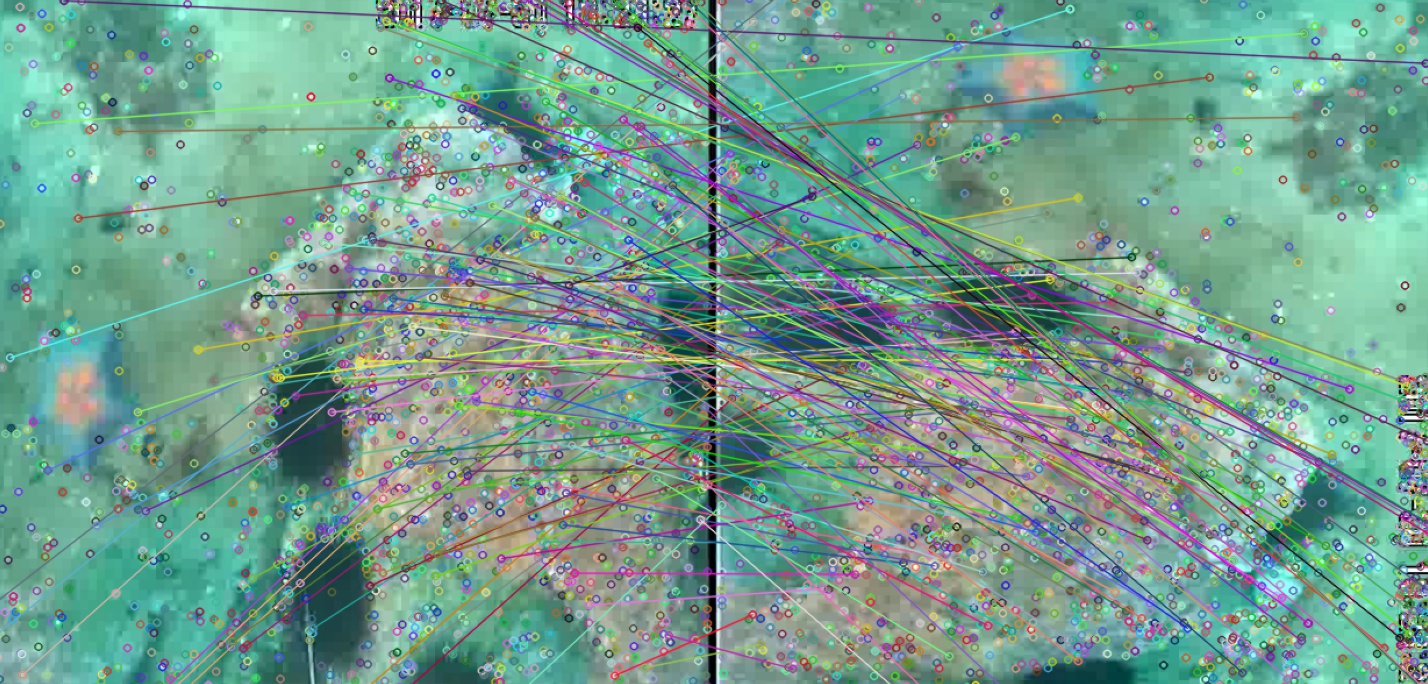}}
	\vspace{3pt}
	\centerline{(b)}
\end{minipage}
\begin{minipage}{0.24\linewidth}
	\vspace{3pt}
	\centerline{\includegraphics[width=\textwidth]{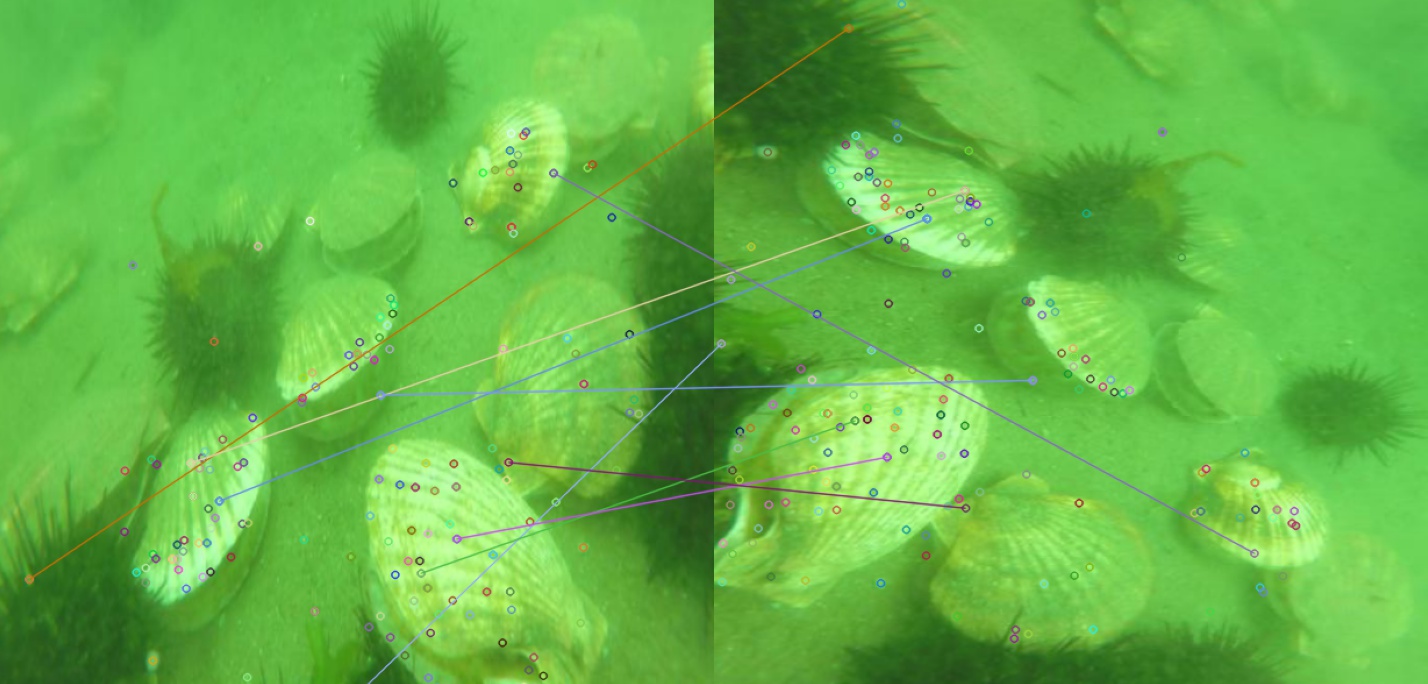}}
	\vspace{3pt}
	\centerline{\includegraphics[width=\textwidth]{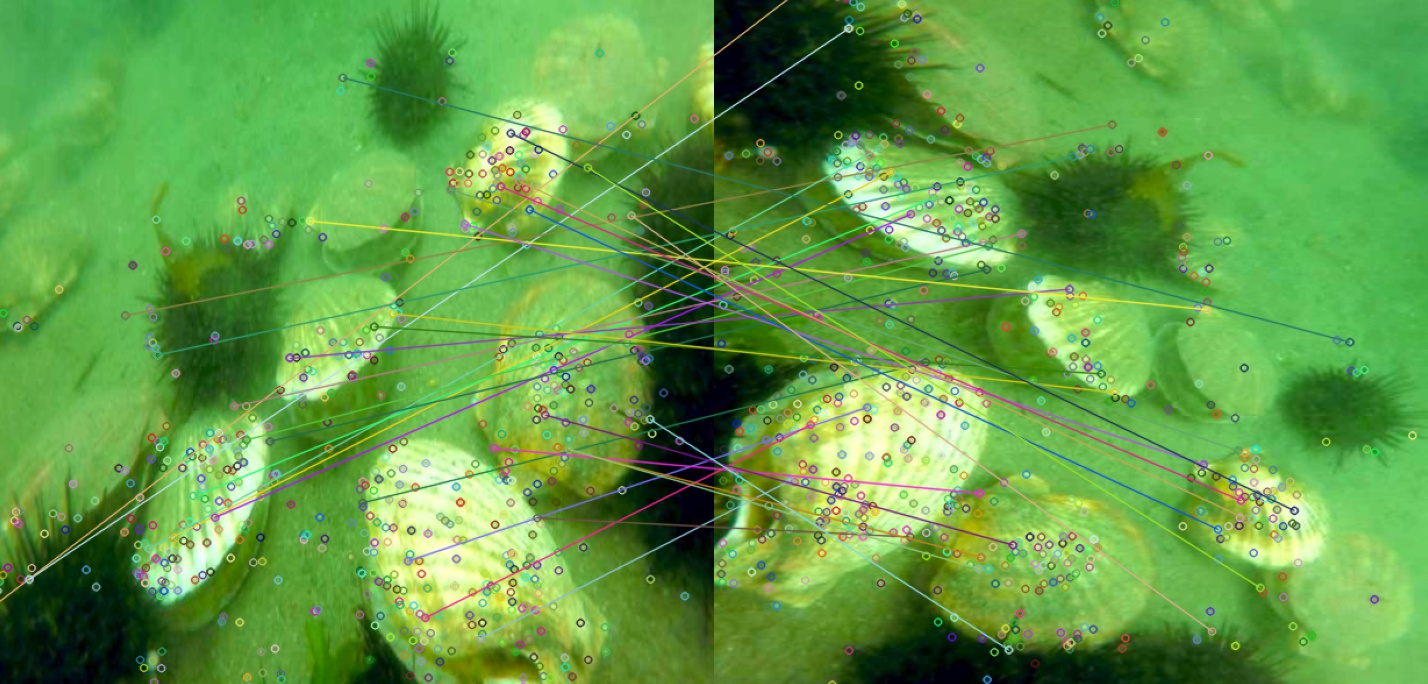}}
	\vspace{3pt}
	\centerline{(c)}
\end{minipage}
\begin{minipage}{0.24\linewidth}
	\vspace{3pt}
	\centerline{\includegraphics[width=\textwidth]{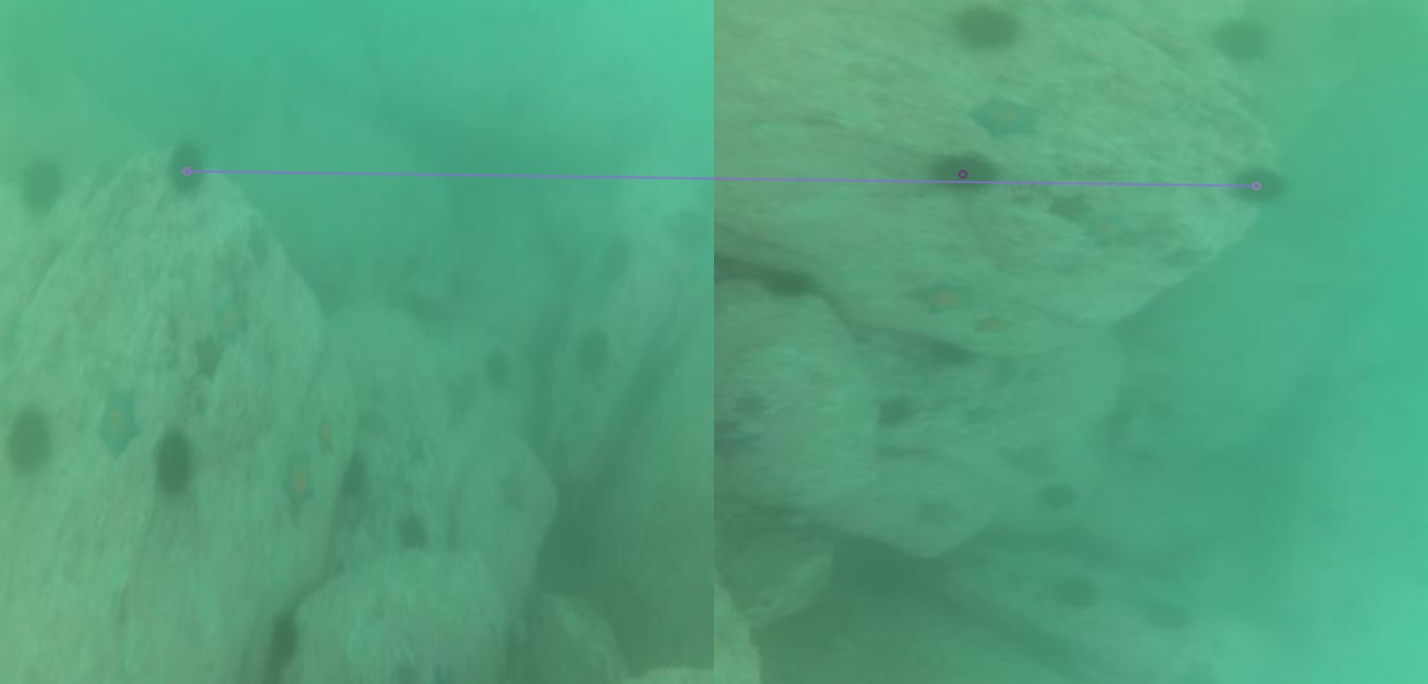}}
	\vspace{3pt}
	\centerline{\includegraphics[width=\textwidth]{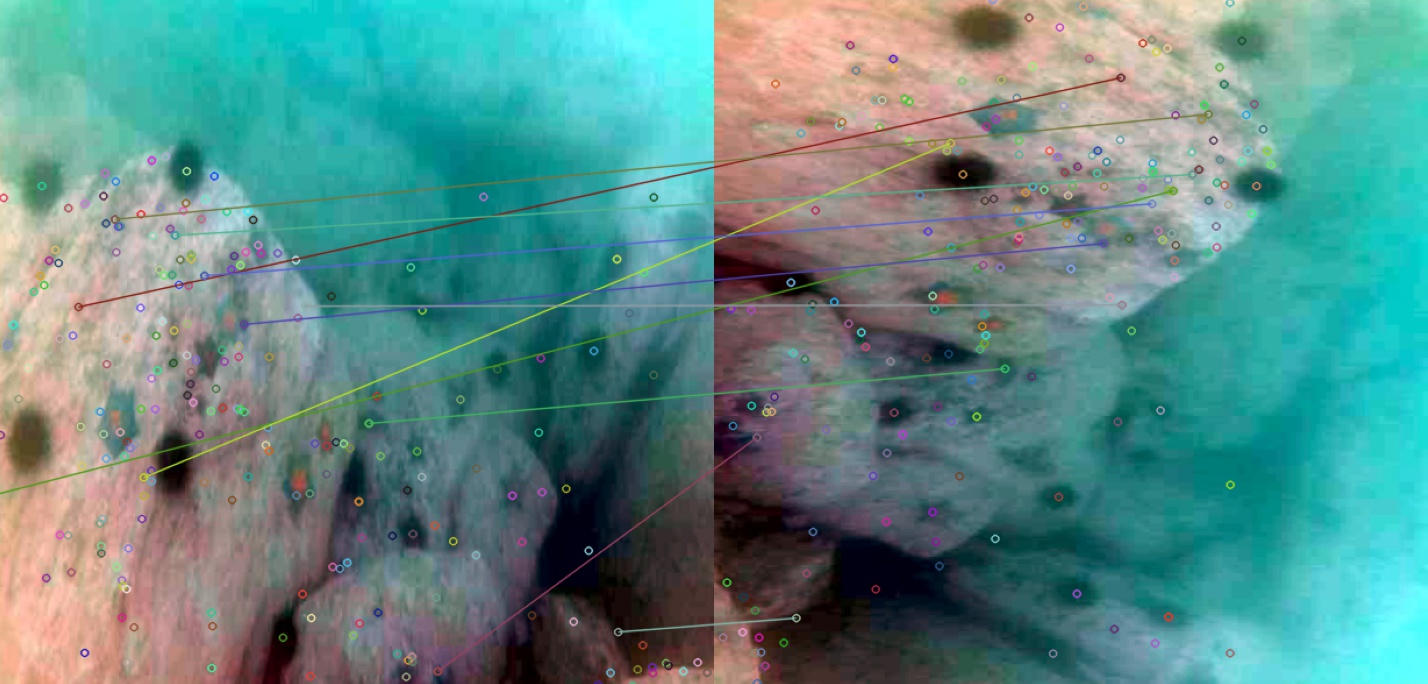}}
	\vspace{3pt}
	\centerline{(d)}
\end{minipage}
\caption{ Results of the SIFT feature matching. The top row is RAWS, and the bottom row is the results of the proposed SAGHS method (a) Low contrast (b) Low resolution (c) Greenish background (d) Bluish background}
\label{sift}
\end{figure*}

Underwater images from various places (low contrast, bluish background, greenish background) are selected to evaluate the efficiency of the proposed SAGHS algorithm. In Figure~\ref{comparise}, it can be clearly seen that the subjective visualization of underwater images after SAGHS processing is better in different water conditions. In low contrast situation, the difference between objects and background contrast in the images after using the SAGHS method is more obvious. In bluish and greenish places, the processed image is supplemented with other colors. At the same time, to further test the profit of the underwater image processing careful way designed in this paper on image point extraction and recognition, and verify the good effects in target detection, this paper takes up the SIFT feature point matching method for experiments. The essence of the SIFT algorithm is to discover key points in the space of the different scales and calculate the direction of the key points. The key points that SIFT gets are those that support out and do not change due to lighting, affine transformations, or noise. The practical idea is to act a certain degree of rotation bias operation on the image group used for comparison, and then act SIFT point matching analysis. It has been discovered that the feature matching algorithm can identify more feature points and produce more accurate matching results for enhanced underwater images. The SIFT result shown in Figure~\ref{sift}.


\subsubsection {Object detection experiments}

\begin{figure*}[!ht]
\centering
 \begin{minipage}{0.45\linewidth}
 	\vspace{3pt}
 	\centerline{\includegraphics[width=\textwidth]{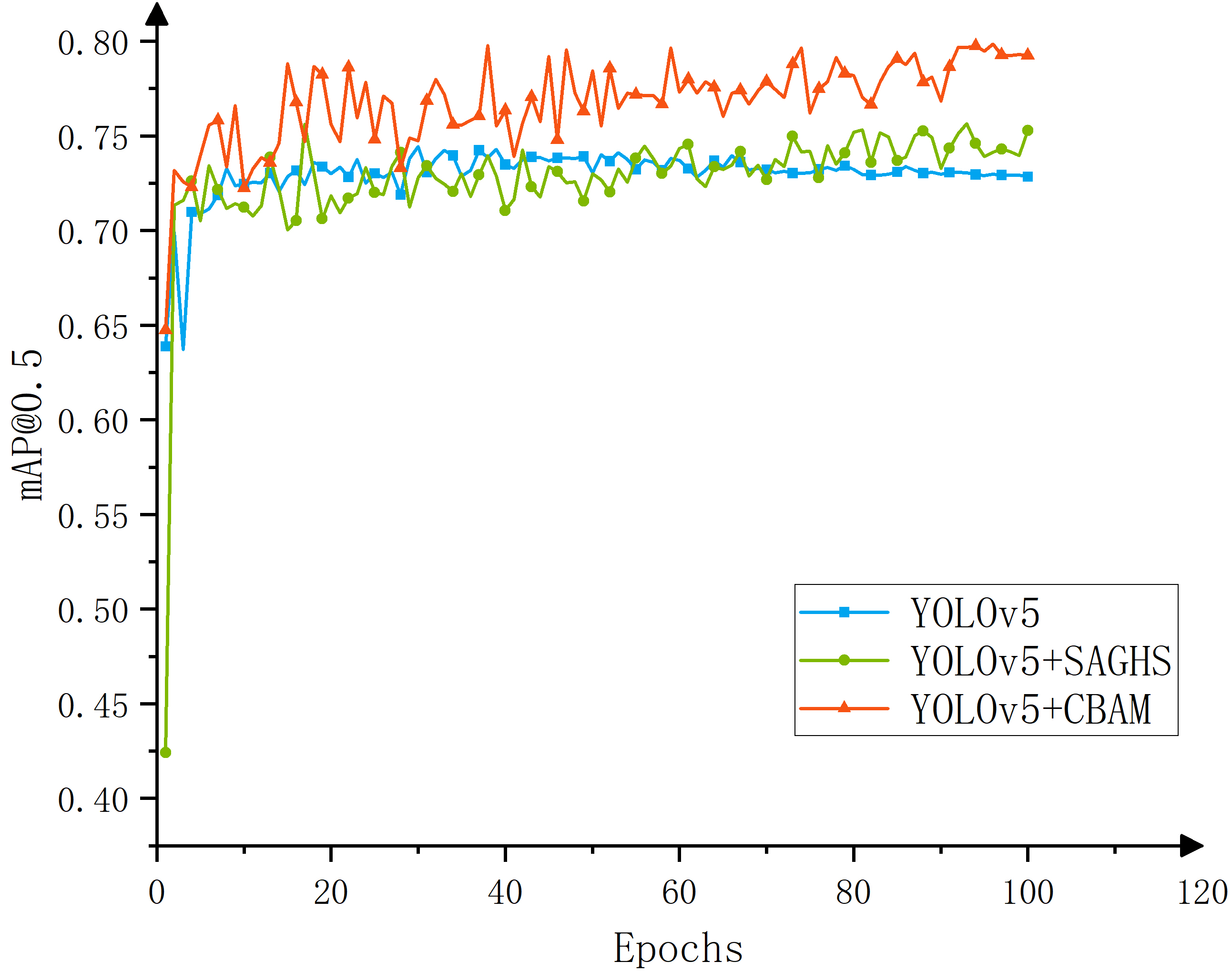}}
 	\vspace{3pt}
  	\centerline{(a)}
 \end{minipage}
 \begin{minipage}{0.45\linewidth}
	\vspace{3pt}
	\centerline{\includegraphics[width=\textwidth]{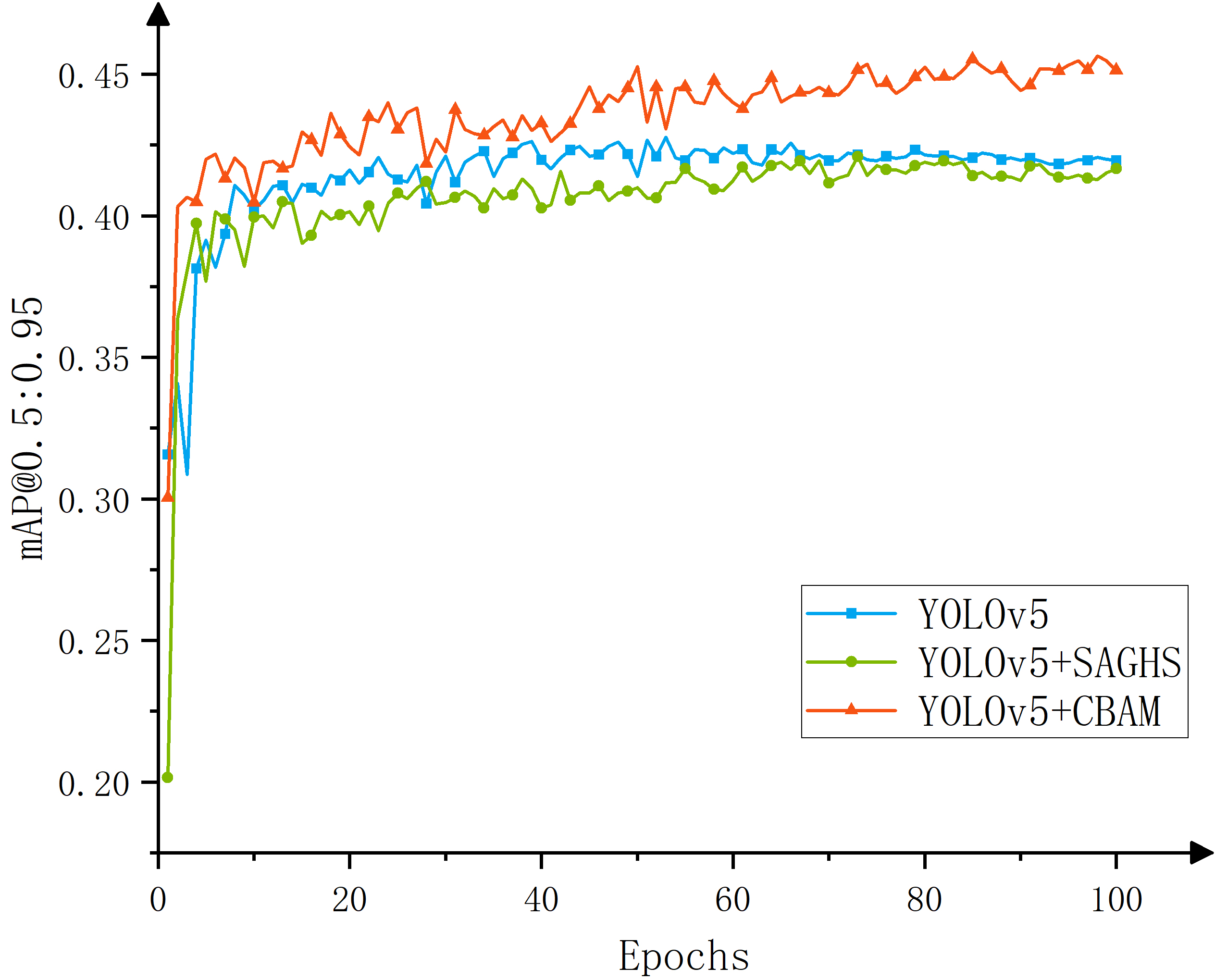}}
	\vspace{3pt}
	\centerline{(b)}
\end{minipage}
\caption{ On the URPC2021 validation set, different algorithms tend to be stable with epoch changes. (a) The detection accuracy in mAP@0.5. (b) The detection accuracy in mAP@0.5:0.95.}
\label{accuracy}
\end{figure*}

In this section, to evaluate the efficacy of the improved YOLOv5 model, it is compared with baseline model.Figure~\ref{accuracy} depicts the mAP curves of three comparison algorithms on the URPC2021 validation set, revealing that the mAP@0.5 (i.e., when IoU is set to 0.5, the AP is calculated for all images in each class, and then averaged over all classes. ) and mAP@0.5:0.95 (i.e., the average mAP across multiple IoU thresholds (from 0.5 to 0.95 with a step size of 0.05) of the three models) tend to be steady as the epochs rise. Each model begins to converge around the fifth epoch. When the mAP@0.5:0.95 was used to evaluate algorithms, the curve of YOLOv5+SAGHS is lower than baseline and YOLOv5+CBAM. This because the underwater image enhancement methods lead to a low precision and the processed image has a little difference between objects and background (such as echinus and rocks), thus effects the detection accuracy. The comparison results of precision, recall and mAP of different algorithms for IoU values of 0.5 and 0.5:0.95 are shown in Table~\ref{table1}. In addition to the comparison with baseline, the mainstream two-stage algorithm Faster RCNN and another model of YOLOv5 were compared. According to Table, it can be concluded that the proposed YOLOv5+CBAM algorithm performs the best, with a score of 79.2\% at the mAP@0.5 and the mAP@0.5:0.95 reaches 45.1\%. When compared to the other methods, the proposed algorithm was improved its mAP@0.5 by roughly 2.4\%-12\%. This improvement is due to the fact that, in terms of structure and processing, the enhanced object detection algorithm and the proposed UIE method are better appropriate for object detection in underwater settings. The Faster RCNN algorithm 
has the worst performance, with 66.4\% on mAP@0.5. YOLOv5+SAGHS outperforms the prior UIE methods in recall metric and solves the low recall problem. The YOLOv5-l model is wider and deeper than the YOLOv5-s model, with a greater emphasis on learning object features. However, YOLOv5-l with a worse score on Recall.

\begin{table}[!htbp]
\caption{Comparison of original and improved YOLOv5 results.}
\centering
\setlength{\tabcolsep}{0.3mm}
\begin{tabular}{ccccc} 
\toprule 
\multicolumn{1}{c}{}&\textbf{Precision}&\textbf{Recall}&\textbf{mAP@0.5}&\textbf{mAP@0.5:0.95} \\  
\hline 
\multicolumn{1}{c}{\textbf{YOLOv5s}}&0.823&0.689&0.729&0.42\\  
\multicolumn{1}{c}{\textbf{YOLOv5l}}&0.811&0.704&0.742&0.427\\
\multicolumn{1}{c}{\textbf{FasterRCNN}}&N/A&N/A&0.664&N/A\\
\multicolumn{1}{c}{\textbf{YOLOv5s+SAGHS}}&0.781$\downarrow$&0.737$\uparrow$&0.753$\uparrow$&0.417$\downarrow$\\
\multicolumn{1}{c}{\textbf{YOLOV5s+CBAM}}&0.837$\uparrow$&0.762$\uparrow$&0.792$\uparrow$&0.451$\uparrow$\\
\bottomrule 
\end{tabular}
\label{table1}
\end{table}

\begin{table}[!htbp] 
\caption{Test results of URPC2021 dataset.(mAP@0.5)}
\centering
\setlength{\tabcolsep}{1.2mm}
\begin{tabular}{ccccc} 
\toprule 
\multicolumn{1}{c}{}&\textbf{holothurian}&\textbf{echinus}&\textbf{scallop}&\textbf{starfish} \\  
\hline 
\multicolumn{1}{c}{\textbf{FasterRCNN}}&0.715&0.855&0.712&0.823\\
\multicolumn{1}{c}{\textbf{YOLOv5s}}&0.685&0.802&0.701&0.753\\  
\multicolumn{1}{c}{\textbf{YOLOv5s+SAGHS}}&0.79&0.918&0.823&0.893\\
\multicolumn{1}{c}{\textbf{YOLOV5s+CBAM}}&0.827&0.934&0.823&0.91\\
\bottomrule 
\end{tabular}
\label{table2}
\end{table}

The morphologies of four types of underwater organisms differ. This research used the mAP@0.5 evaluation index to analyze the differences in detection accuracy between the algorithms for each class of species. Table~\ref{table2} illustrates the detection accuracy of marine organism for each algorithm. The proposed UIE method and detection algorithm perform best in terms of underwater biological identification. For echinus detection, YOLOv5+CBAM has the highest score at 93.4\%. YOLOv5+SAGHS and YOLOv5+CBAM has same score in scallop detection. In complex underwater, the scallop is similar to seabed. This result proves that the SAGHS mechanism effectively separates the object from the background and the CBAM method effectively for small objects difficult to detect. The detection accuracy of YOLOv5+SAGHS and YOLOv5+CBAM for the same organism showed the same general trend.

\begin{figure}[!ht]
  \begin{center}
  \includegraphics[width=3.3in]{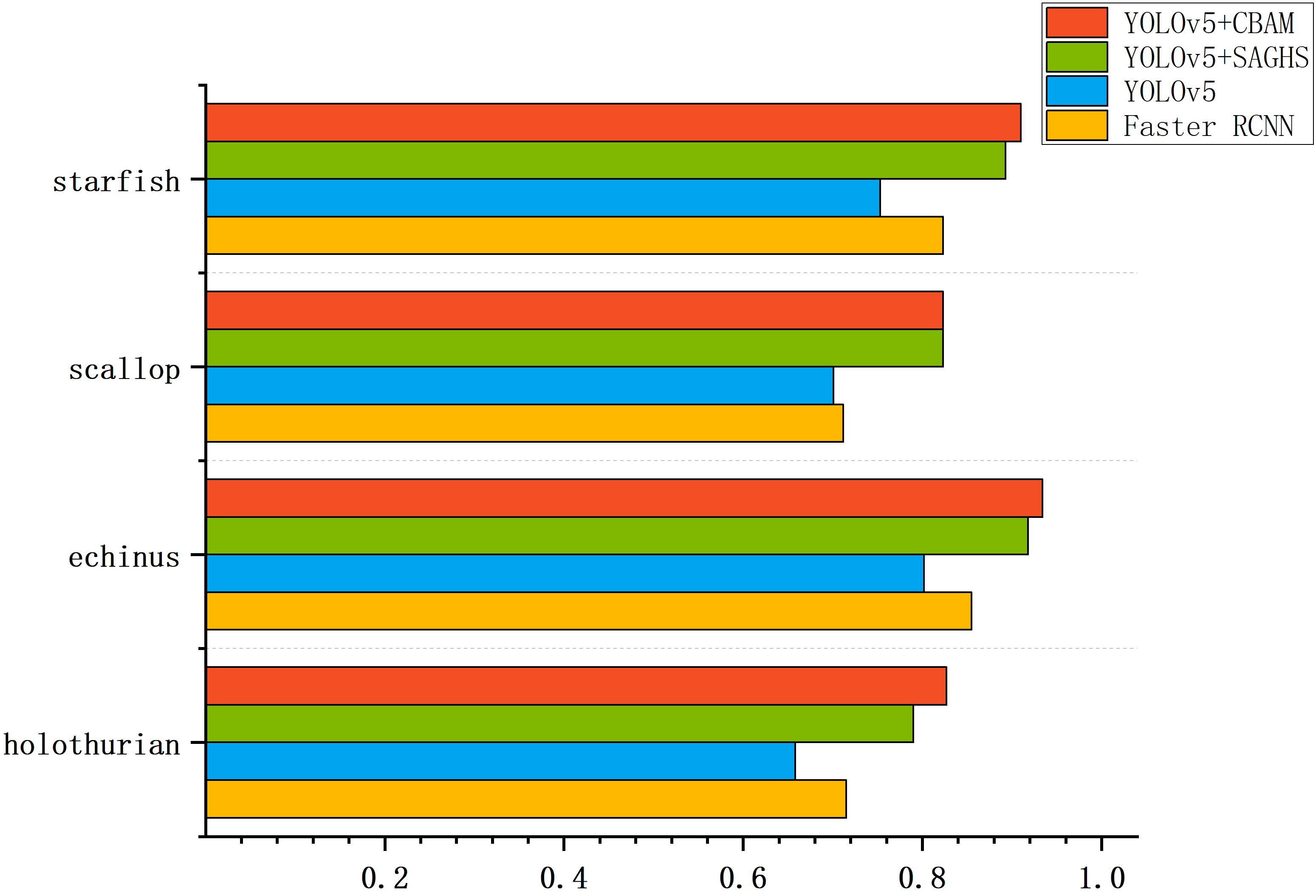}
  \caption{{Clustered bar graph of 4 marine organism species by the four algorithms (The URPC 2021 validation set).}}\label{tiao}
  \end{center}
\end{figure}

Figure~\ref{tiao} shows a clustering histogram comparing the detection accuracy of the four algorithms. It is obvious that the detection of echinus is the most accurate. The holothurian detection has the lowest overall average accuracy. This finding is related to the fact that holothurians marked fewer samples than other types and itself easily deformed when disturbed. Detection of holothurians was significantly improved for detection methods with larger total mAP. Multiple algorithms showed similar trends in recognition accuracy in different marine organisms.

\begin{figure}[!ht]
  \begin{center}
  \includegraphics[width=3.3in]{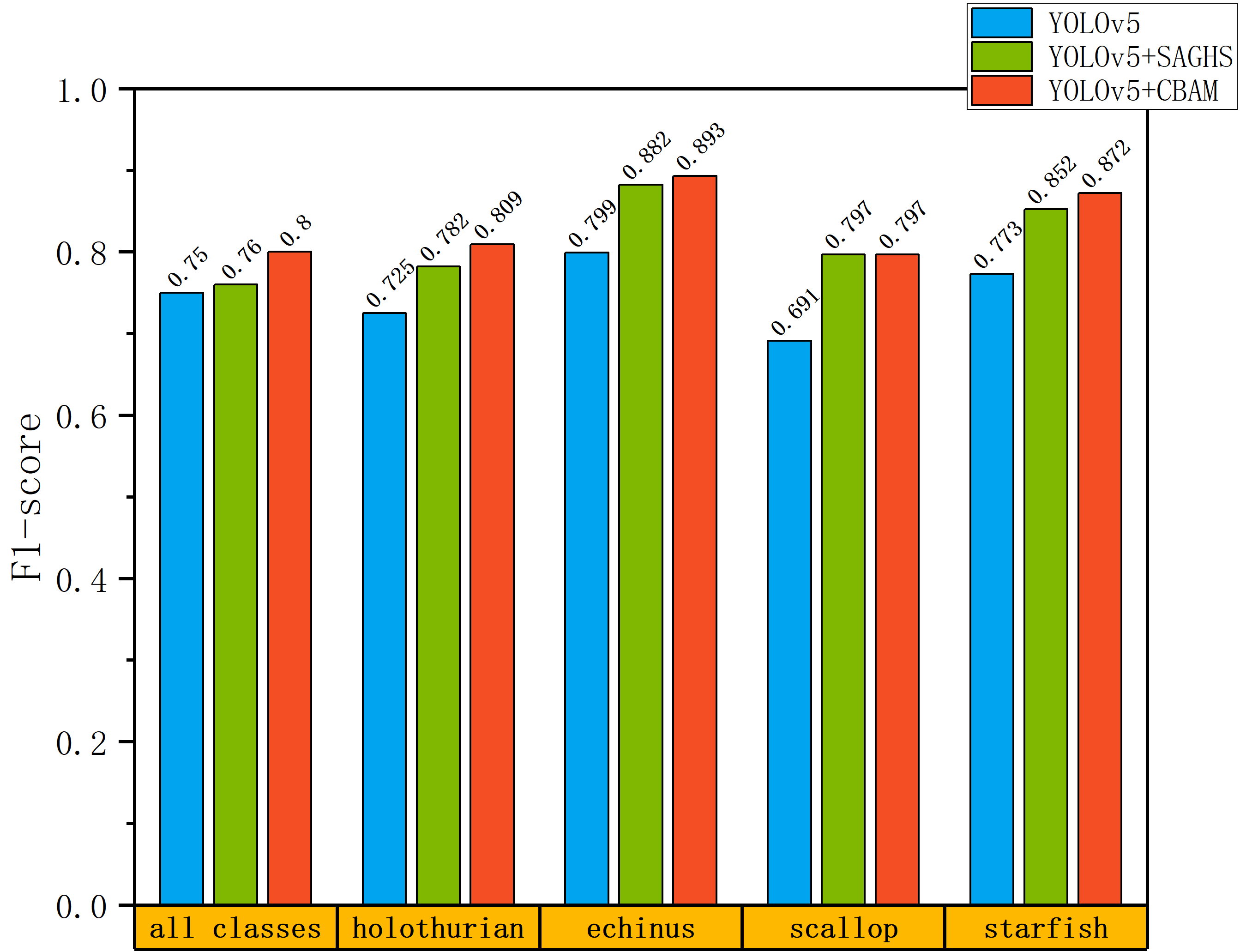}
  \caption{{The all-class comparison histogram between the algorithms of the model in the F1-score evaluation index.}}\label{f1}
  \end{center}
\end{figure}

In addition, F1 scores were also chosen as single and whole category evaluation metrics to validate the detection differences of the three primary comparison algorithms in each category. Figure~\ref{f1} demonstrates that F1 scores may be kept over 0.75 in the majority of categories. The proposed method outperforms the baseline model in various detections and has the greatest overall performance. For other than scallop detection, YOLOv5+CBAM effects is better than YOLOv5+SAGHS. This is because the backbone network with attention mechanism is more capable of extracting meaningful feature information. 

\begin{table}[!htbp] 
\caption{Impact of improvements on network performance.}
\centering
\setlength{\tabcolsep}{1.8mm}
\begin{tabular}{cccc} 
\toprule 
\multicolumn{1}{c}{}&\textbf{Parameters}&\textbf{FPS}&\textbf{Backbone Layer} \\  
\hline 
\multicolumn{1}{c}{\textbf{YOLOv5s}}&7074330&125&283\\  
\multicolumn{1}{c}{\textbf{YOLOv5s+SAGHS}}&7074330&125&283\\
\multicolumn{1}{c}{\textbf{YOLOV5s+CBAM}}&7074940&91&293\\
\bottomrule 
\end{tabular}
\label{table3}
\end{table}

In the inference stage, this paper used the 1400 images of the URPC 2021 test dataset for the model run speed testing. The high FPS value indicates that the model inference is rapid and able to meet real-time requirements. In the same software and hardware environment. The FPS comparison of each model is shown in Table~\ref{table3}. The results of the experiments demonstrate that the FPS of YOLOv5+CBAM has slightly lower from 125 to 91, but the speed still has met with real-time requirements. The image is enhanced before detection, so the FPS of YOLOv5+SAGHS are the same as that of baseline. The backbone network layers and parameters of YOLOv5+CBAM is a minor increase.  
\begin{figure*}[!ht]
	
 \begin{minipage}{0.24\linewidth}
 	\vspace{3pt}
 	\centerline{\includegraphics[width=\textwidth,height=1.5in]{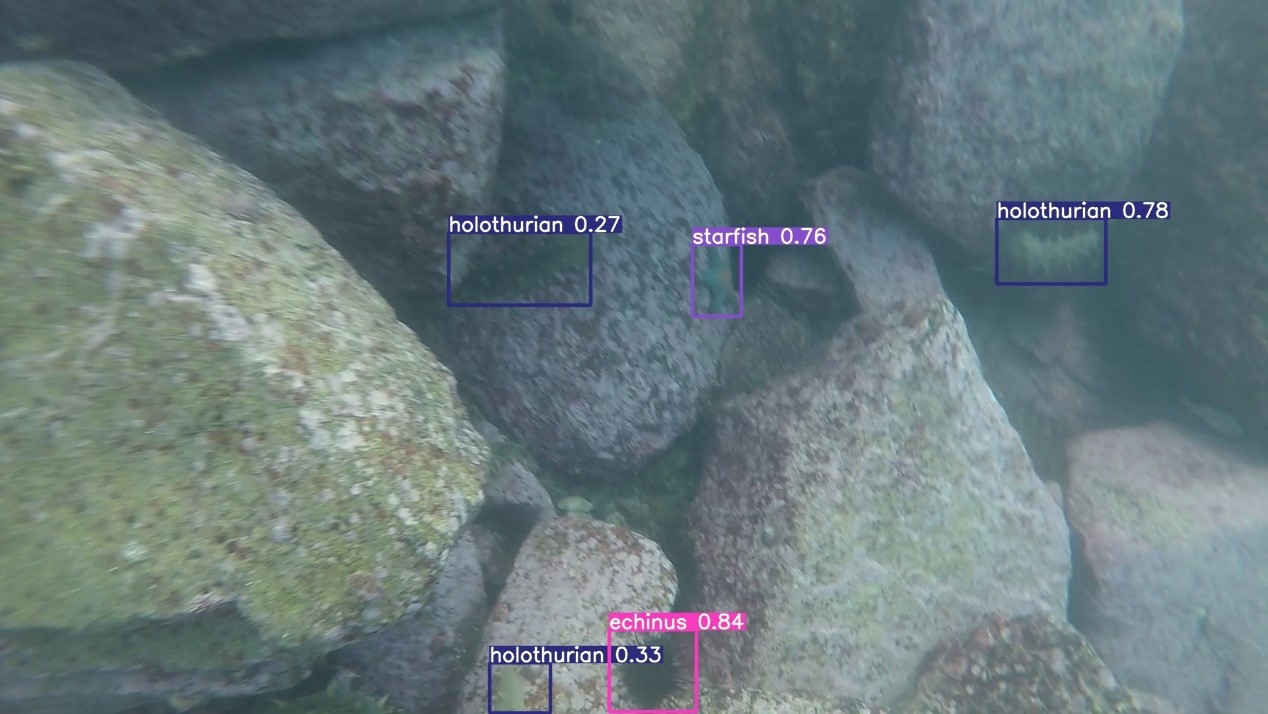}}
 	\vspace{3pt}
 	\centerline{\includegraphics[width=\textwidth,height=1.5in]{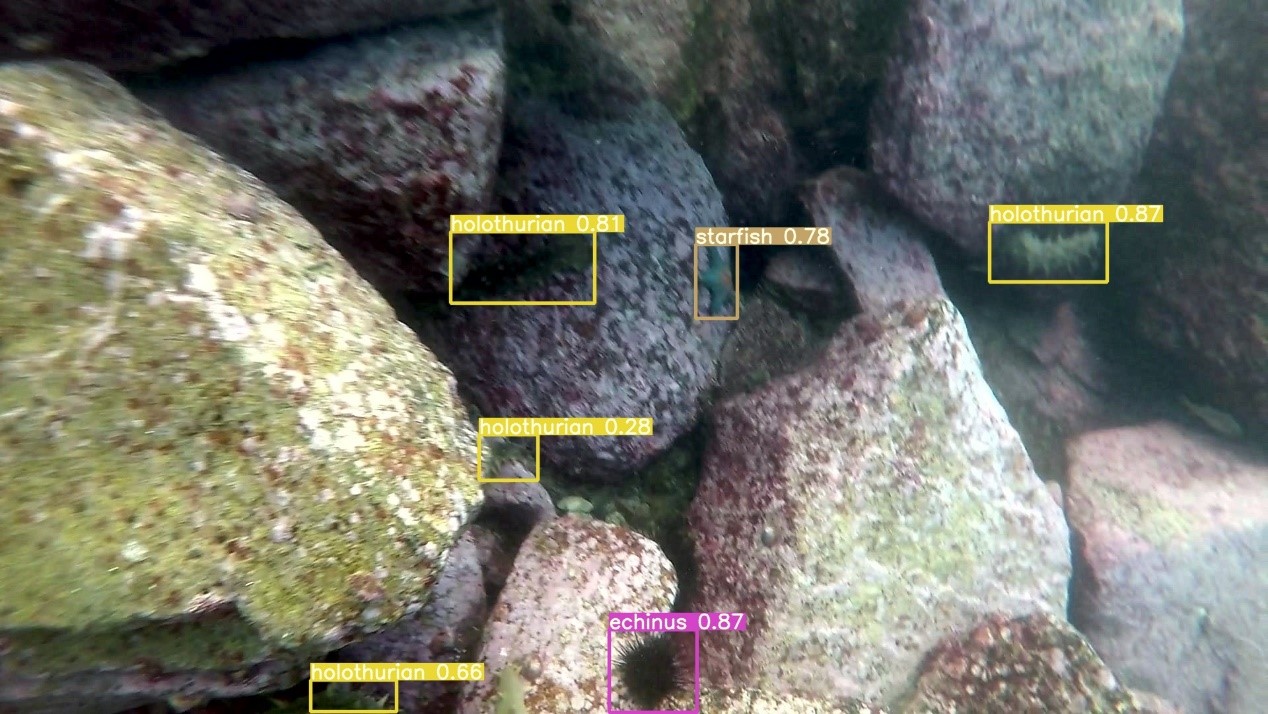}}
 	\vspace{3pt}
 	\centerline{\includegraphics[width=\textwidth,height=1.5in]{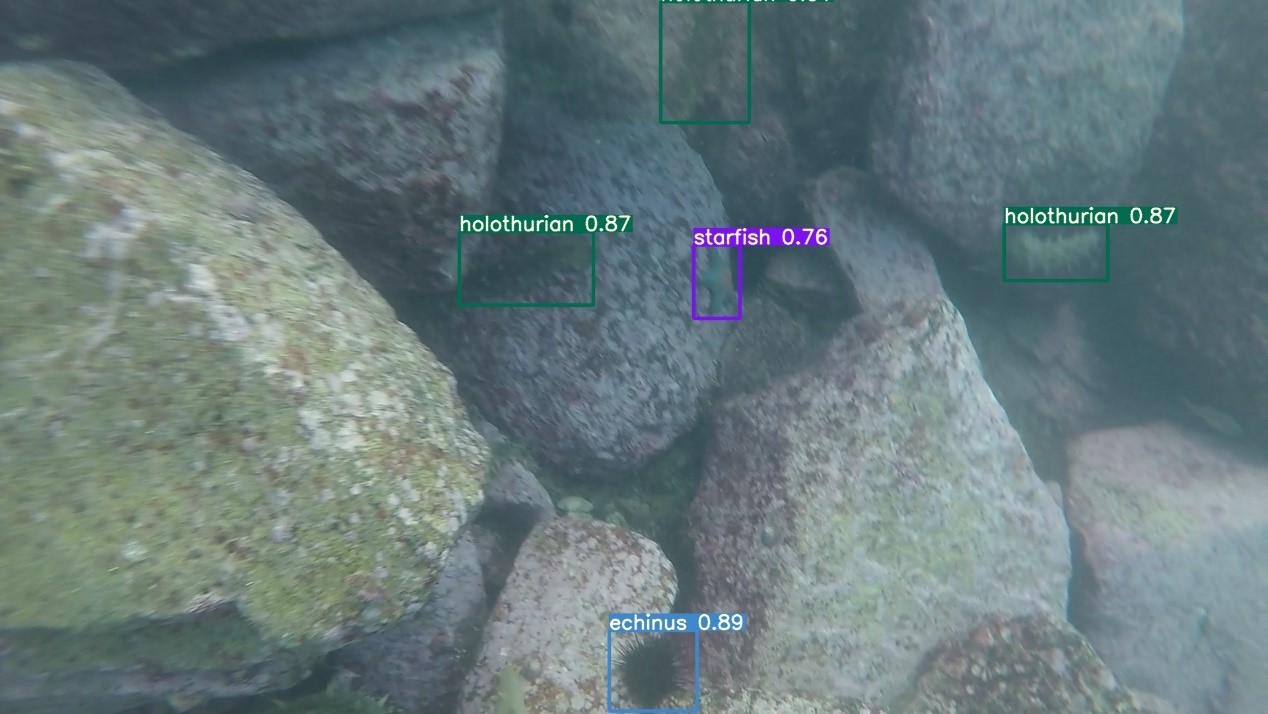}}
 	\vspace{3pt}
 	\centerline{(a)}
 \end{minipage}
 \begin{minipage}{0.24\linewidth}
	\vspace{3pt}
	\centerline{\includegraphics[width=\textwidth,height=1.5in]{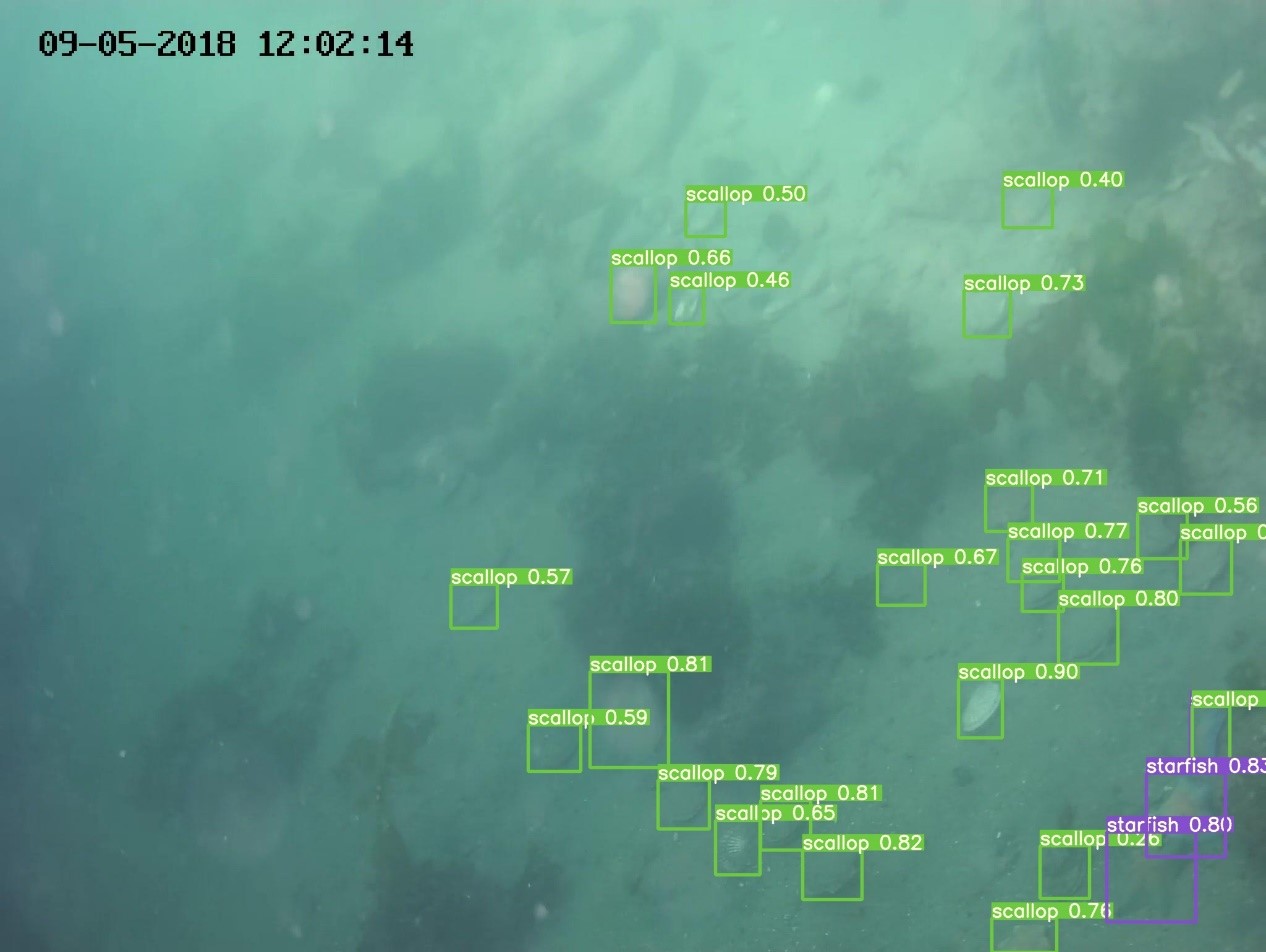}}
	\vspace{3pt}
	\centerline{\includegraphics[width=\textwidth,height=1.5in]{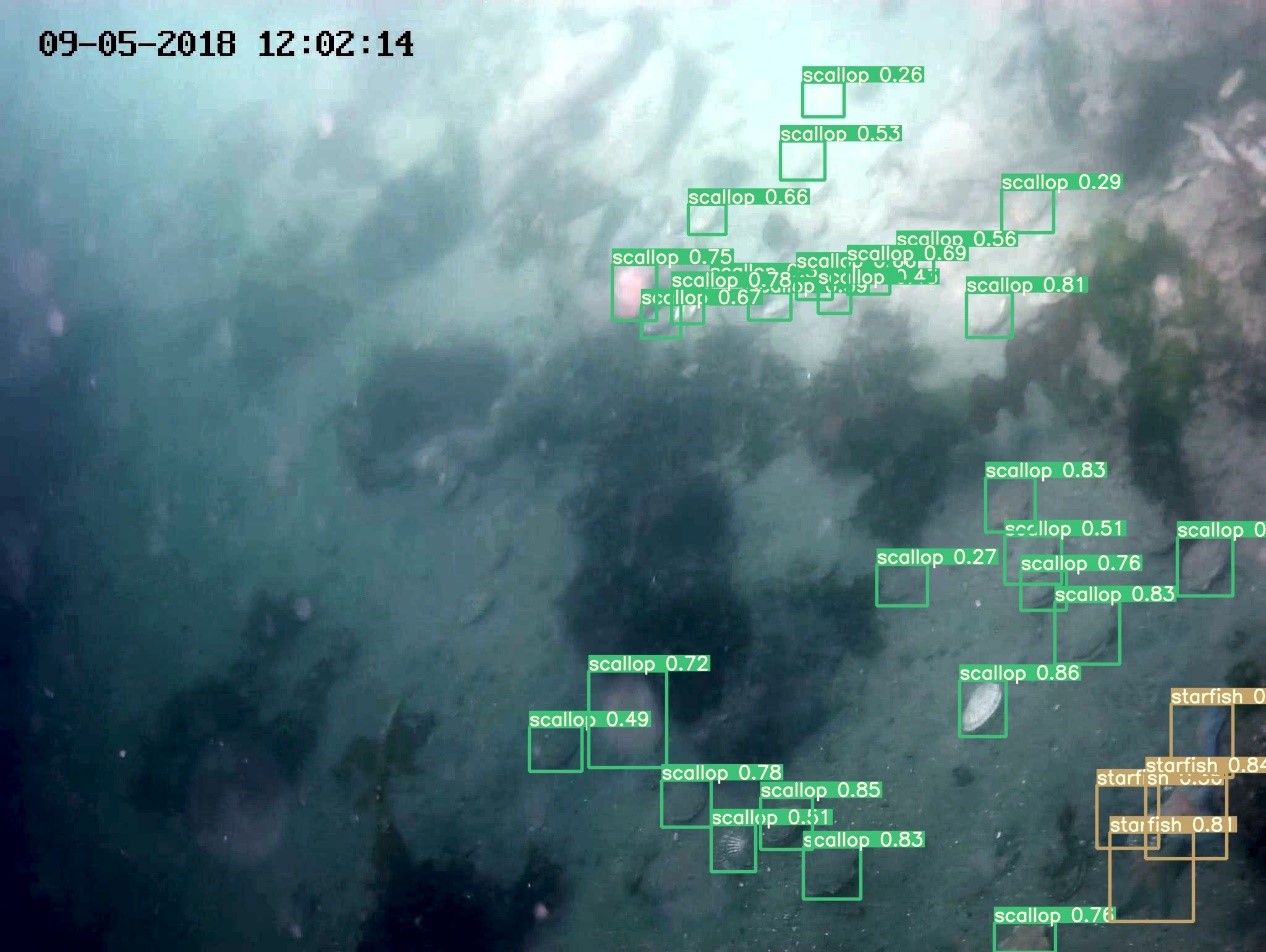}}
	\vspace{3pt}
	\centerline{\includegraphics[width=\textwidth,height=1.5in]{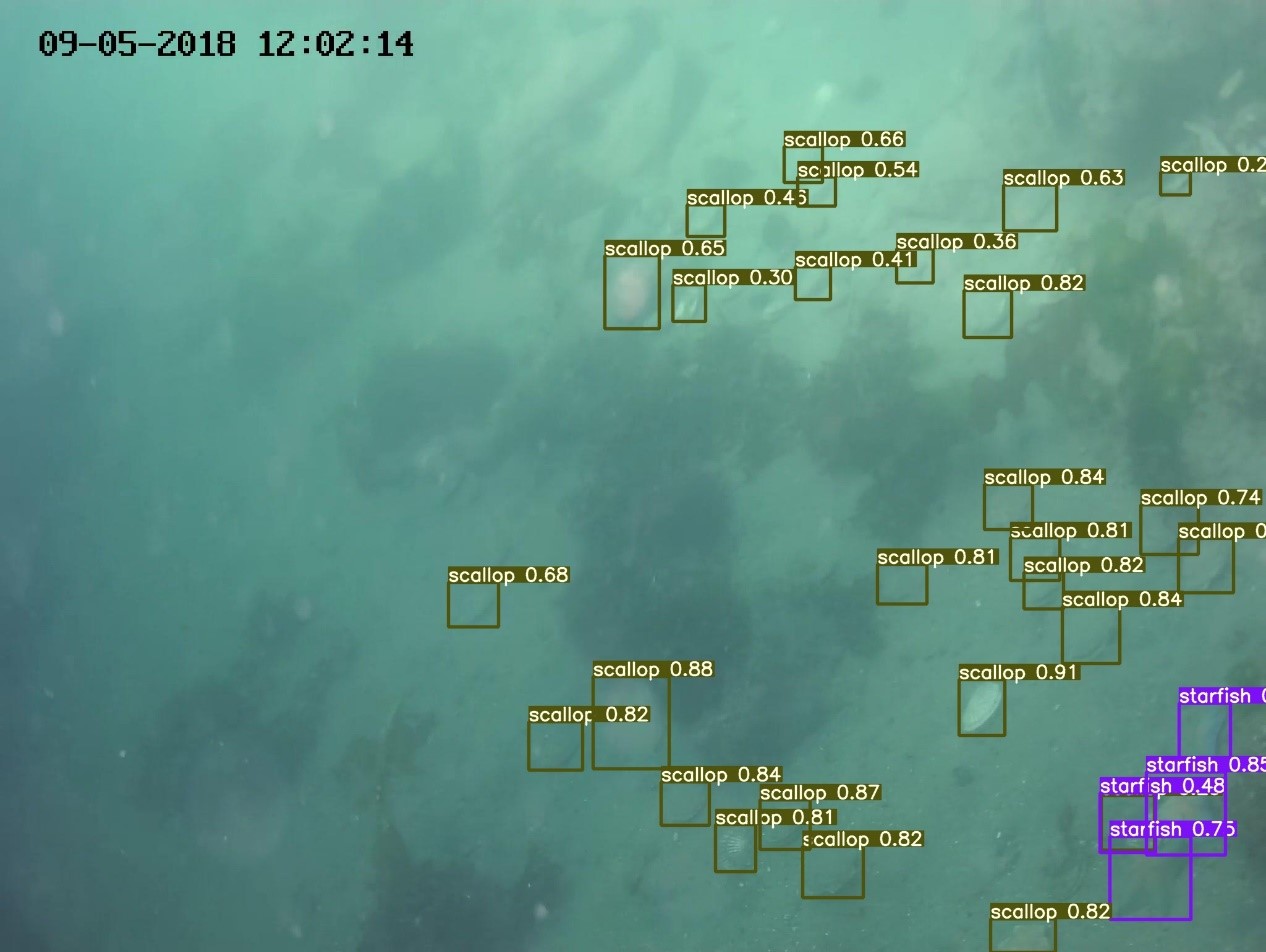}}
	\vspace{3pt}
	\centerline{(b)}
\end{minipage}
\begin{minipage}{0.24\linewidth}
	\vspace{3pt}
	\centerline{\includegraphics[width=\textwidth,height=1.5in]{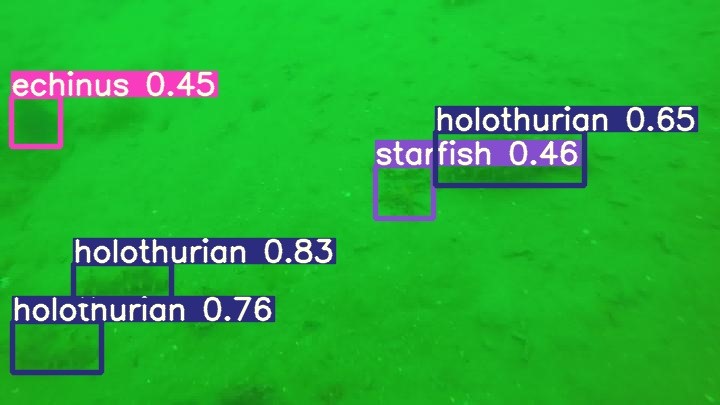}}
	\vspace{3pt}
	\centerline{\includegraphics[width=\textwidth,height=1.5in]{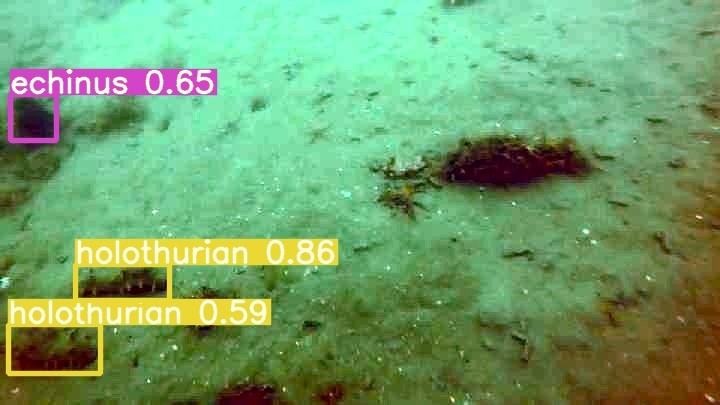}}
	\vspace{3pt}
	\centerline{\includegraphics[width=\textwidth,height=1.5in]{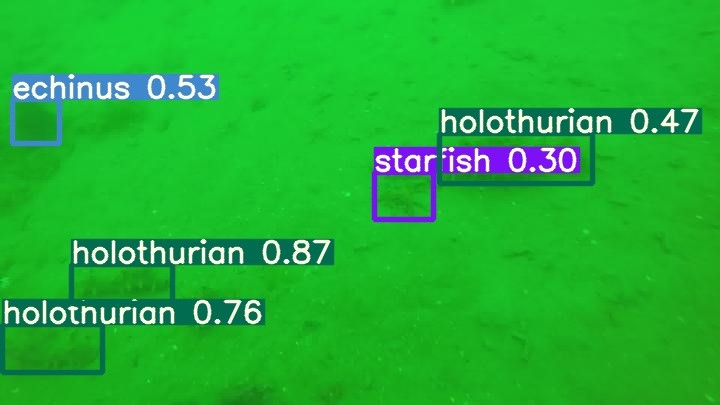}}
	\vspace{3pt}
	\centerline{(c)}
\end{minipage}
\begin{minipage}{0.24\linewidth}
	\vspace{3pt}
	\centerline{\includegraphics[width=\textwidth,height=1.5in]{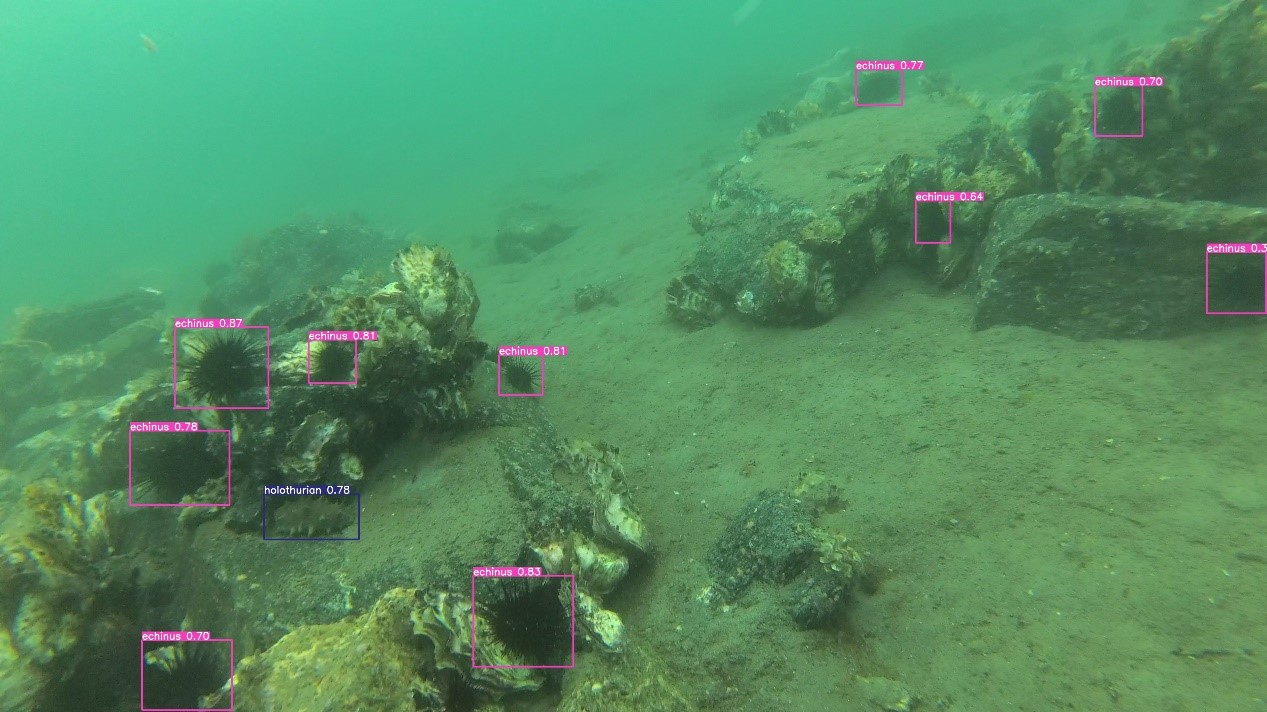}}
	\vspace{3pt}
	\centerline{\includegraphics[width=\textwidth,height=1.5in]{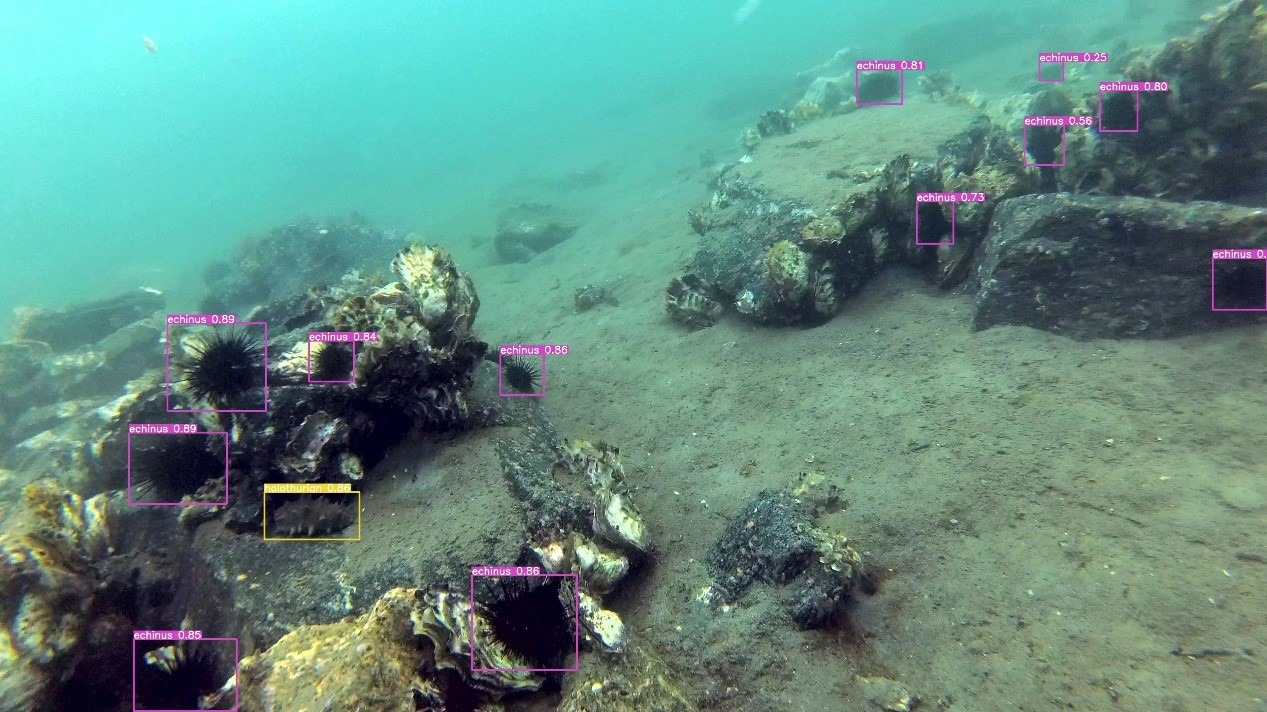}}
	\vspace{3pt}
	\centerline{\includegraphics[width=\textwidth,height=1.5in]{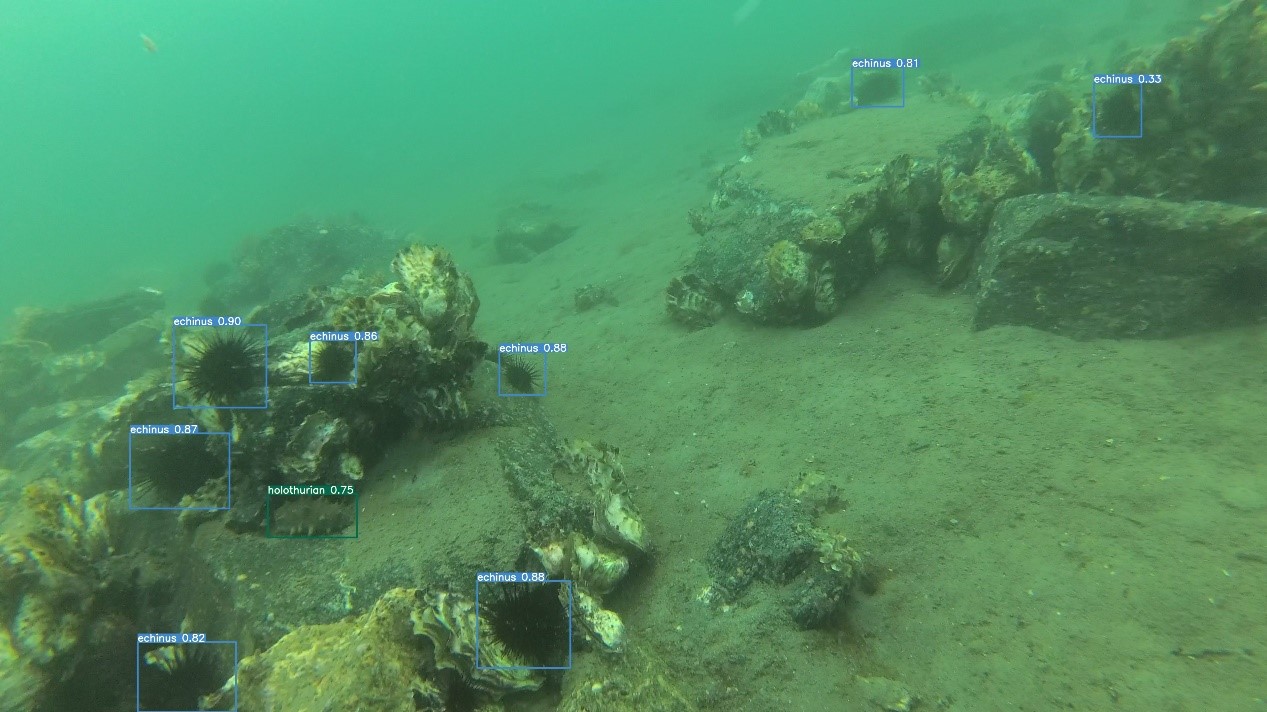}}
	\vspace{3pt}
	\centerline{(d)}
\end{minipage}
\caption{ Comparison of test results. The top row is RAWS, the middle row is the results of the proposed
SAGHS method and the bottom row is the results of improved YOLOv5 with CBAM. (a) Detection of incomplete objects (b) Detection when targets are occluded or overlapped with each other (c) Detection of fuzzy objects and (d) Detection of the object is similar to the background}
\label{last}
\end{figure*}

Figure~\ref{last} presents examples of three models in four frequent scenarios: incomplete objects (Figure~\ref{last} a), occluded and overlapped objects (Figure~\ref{last} b), fuzzy objects (Figure~\ref{last} c) and the objects are similar to the background (Figure~\ref{last} d). From the analysis of the four difficult detection cases in Figure~\ref{last}, the complex underwater environment background and small targets are the main reasons for the missed detection of targets. 
The complex underwater environment background and small objects are the major reasons for missed identification of objects, according to the research of the four challenging detection scenarios in Figure~\ref{last}. In relation to other comparative models, the YOLOv5+SAGHS method aimed at underwater degradation problems to enhance color and contrast quality, which distinguish objects and background so that the target information can be successfully learned, and missed detection can be decreased. In Figure~\ref{last} (a), the detection confidence obtained with YOLOv5+SAGHS are significantly higher when holothurian and rocks are similar in color. The SAGHS methods is used to reduce mistake objects identification in Figure~\ref{last} (b). However, the YOLOv5+SAGHS algorithm can repeatedly and incorrectly detect objects in Figure~\ref{last} (c) (d). Because the scallop and echinus are small and intensive, the YOLOv5+CBAM algorithm was used to detect them. The CBAM mechanism not only can effectively improve the backbone network’s capacity to extract objects feature information but also make it easier to recognize marine organisms in different underwater scenes. A lightweight attention mechanism combined with a backbone network fusion strategy can increase the generalization performance of the network model.
\section {Conclusion}

This paper proposed a self-adaptive global histogram stretching and an improved YOLOv5 underwater organism detection algorithm to tackle the problems of underwater degradation and low detection accuracy. Originally, a histogram adaptive extending range approach was devised for the underwater image enhancement algorithm to improve visibility and details while minimizing artifacts and noise. In addition, for recognizing the small and overlapping objects caused by underwater biological characteristics, an advanced YOLOv5-based underwater object detection algorithm in conjunction with the CBAM model is designed. The experimental results reveal that, when compared to the previous or superior object algorithm, the proposed algorithm provides significantly enhanced detection accuracy, demonstrating the algorithm's usefulness. The concept of attention mechanism and image enhancement is used to the learning of complicated underwater environment and biological features information in this article, which benefits the accuracy of underwater object detection and has particular practical application effects. In the future, the network model can be further optimized to be lighter while ensuring detection accuracy. The algorithm is applicable in different disciplines (e.g., mariculture resource surveys, underwater operational robots, and other underwater image and video object detection).

\section*{Acknowledgments}
This work was supported in part by the China Postdoctoral Science Foundation 2020M670778; the Natural Science Foundation of Liaoning Province 2021-BS-051; the Northeastern University Postdoctoral Research Fund 20200308; the Scientific Research Foundation of Liaoning Provincial Education Department LT2020002; the National Natural Science Foundation of China under Grants U2013216, 61973093, 61901098, 61971118, and 61973063; the Fundamental Research Funds for the Central Universities N2026006, N2011001 and N2211004; the China Scholarship Council; the Liaoning Key R\&D Project 2020JH2/1010\\0040.



\bibliographystyle{unsrt}

\bibliography{myref}

\bio{}
\endbio

\endbio

\end{document}